\documentclass[article,onecolumn]{IEEEtran}

\AtBeginDocument{%
  \providecommand\BibTeX{{%
    \normalfont B\kern-0.5em{\scshape i\kern-0.25em b}\kern-0.8em\TeX}}}

\usepackage{amsmath,amssymb,amsfonts}
\usepackage{algorithmic}
\usepackage{graphicx}
\usepackage{textcomp}
\usepackage{xcolor}
\usepackage[normalem]{ulem}
\usepackage[ruled,linesnumbered]{algorithm2e}
\usepackage{amsmath}
\usepackage{epsfig, subfigure, amsmath, amssymb, wrapfig}

\newcommand{\name}{{\textsc{\small{REPlanner}}}\xspace}    

\definecolor{aqua}{rgb}{0.7,0,0.7}

\makeatletter
\let\@authorsaddresses\@empty
\makeatother

\begin{document}

\title{Efficient UAV Trajectory-Planning using Economic Reinforcement Learning}

\author{
\IEEEauthorblockN{Alvi Ataur Khalil\IEEEauthorrefmark{1}, Alexander J Byrne\IEEEauthorrefmark{1}, \\Mohammad Ashiqur Rahman, and Mohammad Hossein Manshaei}
\IEEEauthorblockA{
\\Analytics for Cyber Defense (ACyD) Lab \\
Florida International University, Miami, USA\\
	\{akhal042, abyrn010, marahman, mmanshae\}@fiu.edu
	}
}

\maketitle
\begingroup\renewcommand\thefootnote{\IEEEauthorrefmark{1}}
\footnotetext{Khalil and Byrne are the co-first authors of this paper.
}

\begin{abstract}
Advances in unmanned aerial vehicle (UAV) design have opened up applications as varied as surveillance, firefighting, cellular networks, and delivery applications. Additionally, due to decreases in cost, systems employing fleets of UAVs  have become popular. The uniqueness of UAVs in systems creates a novel set of trajectory or path planning and coordination problems. Environments include many more points of interest (POIs) than UAVs, with obstacles and no-fly zones. 
We introduce \name{}, a novel multi-agent reinforcement learning algorithm inspired by economic transactions to distribute tasks between UAVs. This system revolves around an economic theory, in particular an auction mechanism where UAVs trade assigned POIs.
We formulate the path planning problem as a multi-agent economic game, where agents can cooperate and compete for resources. We then translate the problem into a Partially Observable Markov decision process (POMDP), which is solved using a reinforcement learning (RL) model deployed on each agent. 
As the system computes task distributions via UAV cooperation, 
it is highly resilient to any change in the swarm size. Our proposed network and economic game architecture can effectively coordinate the swarm as an emergent phenomenon while maintaining the swarm's operation. 
Evaluation results prove that \name{} efficiently outperforms conventional RL-based trajectory search.

\end{abstract}

\section{Introduction}
\label{sec:introduction}
Unmanned aerial vehicles (UAVs) are applicable to a wide-ranging set of problems such as fire fighting, security monitoring, agriculture, edge computing, 3D mapping,  and network support~\cite{shakhatreh2019unmanned}. Fire fighting problems center around tracking and finding fires, whereas security applications focus on monitoring and finding targets. On the other hand, agricultural problems center around field monitoring and data harvesting, while edge computing and network support are focused on data harvesting and load reaction. All of these problems can be abstracted to a set of partially observed points and must be traveled to in the shortest amount of time possible, and then some task must be carried out in the vicinity of this point. 
Swarm surveillance missions are essential in both civilian and military contexts, where solutions must be secure, reliable, and efficient. The problem is computationally expensive as solutions must be provided for each UAV and take into account the paths of other UAVs to avoid collisions and use resources efficiently. Several UAVs tracking a low priority target while ignoring higher priority targets is not an efficient use of resources regardless of how optimized the individual paths are.  Current methods include Swarm intelligence optimization (SIO), formal methods, convex optimization, and graph-based methods~\cite{duan2014pigeon,liu2017survey,naazare2019application}. Besides SIO, these methods suffer from two glaring issues. They require a static environment model where each agent is destined for some fixed trajectories and become very computationally expensive as the system’s complexity increases. Further, all of these methods consist of UAVs taking commands from a single point, which, if compromised, brings the entire system down. Thus, a framework is needed that considers the UAV swarm’s characteristics from the ground up to provide a more secure and reliable service.

The cutting edge of UAV swarm technology is reinforcement learning (RL). The problem is typically modeled as a partially observable markov decision process (POMDP). This means our state is an incomplete representation of the environment, and so our agents do not and will not know the full effects of their actions. Nevertheless, our agents attempt to learn the strategy that results in the greatest reward. With a well designed reward function, agents can learn to act effectively based solely on sensor input. These methods are able to react to a changing environment in real time, and decentralized allowing continued operation when a subset of members fails. Where traditional methods must recalculate with changes in the environment, and completely fail with reassigning unknown targets, RL methods adapt paths immediately. 
In the UAV swarm domain, RL is used to control agents for proving celluar internet~\cite{hu2020reinforcement} and data harvesting~\cite{bayerlein2020multi} missions while avoiding collisions~\cite{wang2020two}. Each UAV is controlled by its own agent, acting in it’s own interest, and attempting to gain the most reward for itself. When attempting to cooperate, mission time increases in~\cite{wang2020two}. The agents in~\cite{bayerlein2020multi} left out a significant portion (20\%) of data points as the environment grew in size. In~\cite{walker2020multi}, significant portions of the environment were left unexplored. 

While all of these papers frame the problem as a cooperative one, it is immediately seen that agents are also competing for reward, as an agent completing a shared point, effectively removes the opportunity for another agent to gain that reward and potentially wastes fuel. It is natural to then frame the problem in terms of game theory, and agent to agent communications as part of an economic game. This framing allows us to anticipate individual maladaptive strategies that may be employed by members of the swarm, such as intentional collisions. Further this framing enables agents to specialize, taking on different roles to better enable mission completion, allowing us to optimize the proportion of agents following different policies, i.e., exploration, following, and monitoring. 
\name{} agents enter into bids for actions, allowing the swarm as a whole to find the value of actions through pricing. As this emerges from the interactions of agents, it is decentralized and comes at low computational cost. Further as we employ RL, the agent is able to learn which strategies are effective in the context of other agents' strategies in addition to current configuration of the environment. Since pricing is a proxy to the swarm’s valuation, it provides a quick way of monitoring the behaviour of the swarm as a whole, helping to deal with the issue of opaqueness in machine learning. As a result, \name{} allows for dynamic optimization of strategies and paths by individual agents, while keeping computational costs low, and allowing easy manipulation of the swarm’s priorities. We tested differences between standard q-learning and our economic variant finding up to a 33\% increase in path distance efficiency, an 18\% decrease in time to completion, and a 200\% increase in reward gain. This reward gain is particularly telling, as it shows our model adapts to it's directive to act in the environment far better than standard Q-learning. With fine tuning of the reward function's design we can expect even better performance on other metrics. Clearly validating the basis of our model and consequent framework, our contributions are as follows:
\begin{itemize}
    \item We design and implement \name{}, a novel distributed optimization technique, which mimics economic exchange to re-assign objects to agents.
    \item We introduce an auction mechanism for distributed trajectory or path planning system which anticipates competitive as well as cooperative strategies from agents. 
    \item Our framework shows intelligent choice of trajectory which converges faster than traditional Q-learning.
    \item We evaluate the proposed \name{} framework with respect to time, distance, and goal completion. The evaluation results show that our proposed framework was more efficient than standard Q-learning.
\end{itemize}

\noindent\textit{Organization:} The rest of the paper is organized as follows: We discuss sufficient preliminary information in Section~\ref{sec:background}. The related works are discussed in Section~\ref{sec:related_works}. We introduce our proposed \name{} framework in Section~\ref{sec:proposed_framework}. In Section~\ref{sec:technical_details}, we discuss the technical details of the frameworks and the complete analysis of our algorithms. In Section~\ref{sec:evaluation_setup}, we explain the evaluation setup and dataset. The empirical analysis and findings are formulated in Section~\ref{sec:evaluation_result}. At last, we conclude the paper in Section~\ref{sec:conclusion}.

\section{Background}
\label{sec:background}

\begin{figure}[t]
    \centering
    \includegraphics[width=0.5\columnwidth]{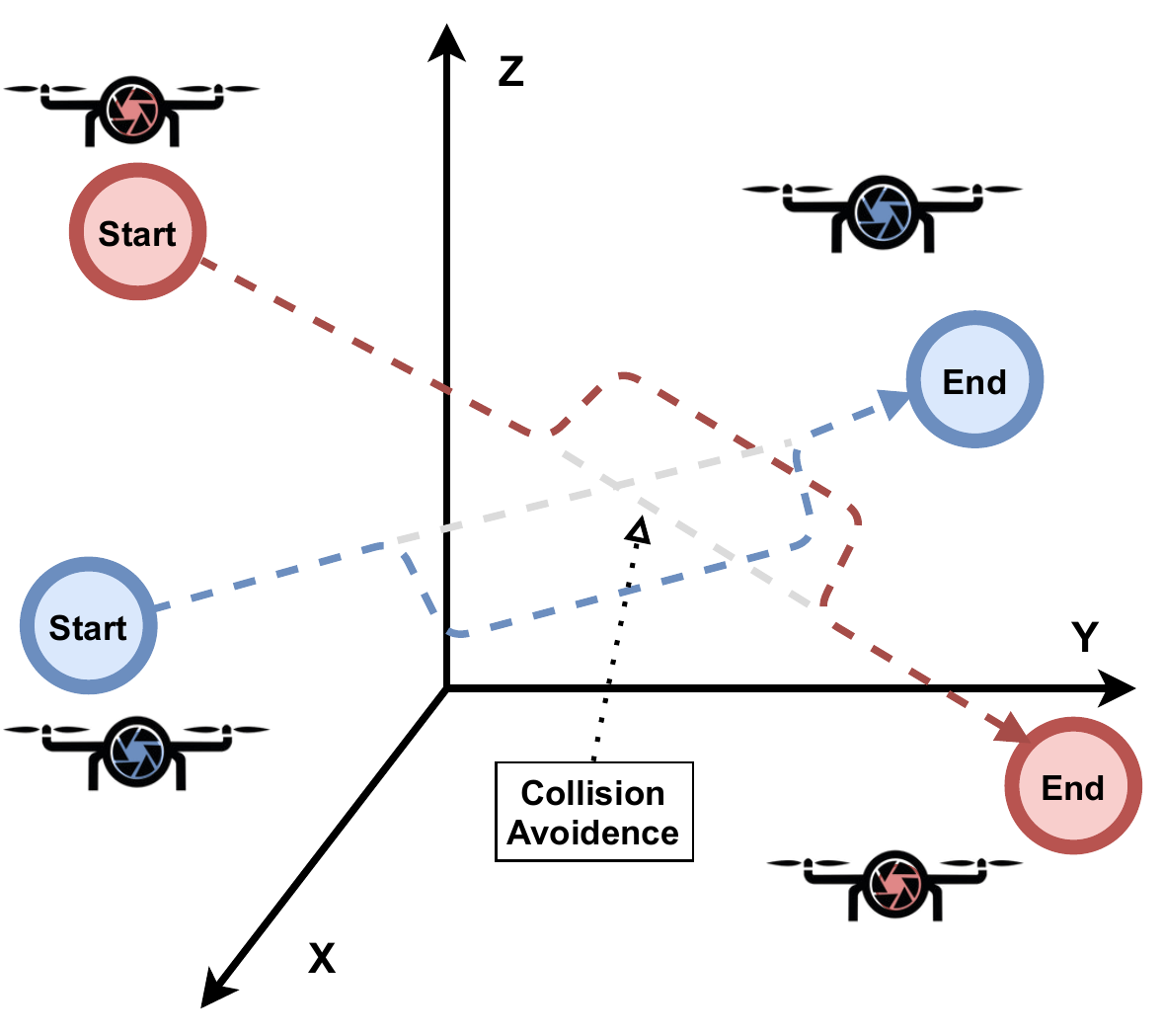}
    \vspace{-9pt}
    \caption{UAV path-planning}
    \label{fig:pp}
    \vspace{-6pt}
\end{figure}
In this section,  we discuss the necessary preliminary information that helps to explain the problem domain and proposed \name{} mechanism.

\subsection{UAV path-planing}
Path planning is referred to the mechanism of finding an optimal path between source and destination and it is one of the most important problems to be explored in the unmanned aerial vehicles (UAVs) domain. The main objective of UAV path planning is to design a cost effective flight path that meets the UAV performance requirements with minimal probability of being destroyed during the flight~\cite{aggarwal2020path}. The aim of UAV path planning also includes providing a collision-free environment to the UAVs, as presented in Fig.~\ref{fig:pp}. UAVs path planning generally includes three key terms~\cite{giesbrecht2004global}: Motion planning, trajectory planning, and navigation. Motion planning satisfies constraints like flight path, turning a crank in the motion of the path planning. On the other hand, trajectory planning encloses the path planning having time, velocity, and kinematics of UAVs motion whereas navigation is concerned with collision avoidance, and localization. For UAV path planning, a three-dimensional environment is required, as simple two-dimensional path planning methods will not be able to find the obstacles and objects in a complex environment. There are numerous path-planning methods for UAVs to navigate in a obstacles filled three-dimensional domain that formulate path-planning as an optimization problem. The algorithms that are used to solve these optimization problems can be precisely described by following groups~\cite{goerzen2010survey}: heuristic and non-heuristic (or exact) methods. Heuristic
methods generally trade the optimality of the solution for
computational time-efficiency whereas, non-heuristic methods use computationally expensive mathematical principles for obtaining
the solutions. Most of the approaches start by breaking down the detected area or map into computational domains which helps in the generation of possible UAV trajectories.


\begin{figure}[!t]
    \centering
    \includegraphics[width=0.45\columnwidth]{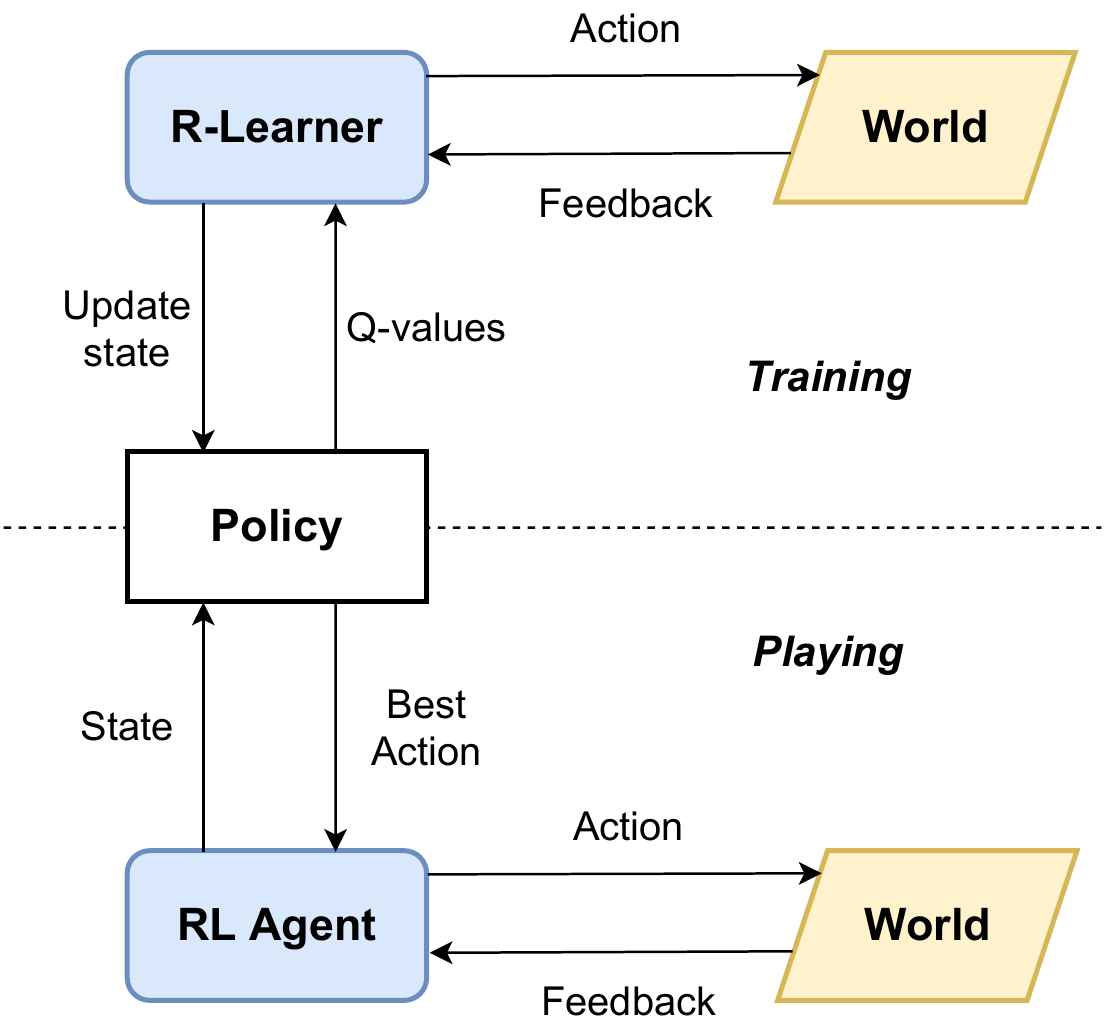}
    \vspace{-9pt}
    \caption{Reinforcement Learning Architecture}
    \label{fig:rl}
    \vspace{-12pt}
\end{figure}

\subsection{Reinforcement learning}

Reinforcement learning refers to the field of Machine learning that deals with the mapping of situations to actions in order to maximize the achievement of the actor. In RL, an agent interacts with its environment to learn an optimal policy which maximizes expected cumulative rewards for a given task~\cite{henderson2017deep}. The objective of RL is to maximize future rewards. However, since an actor cannot predict the future changes in its environment perfectly, value functions reflect the actor’s empirical estimates for its future rewards. The distant future rewards are often discounted temporally so that more immediate rewards exert stronger influence on the actor’s behavior. The RL theory utilizes two different types of value functions, action value function and state value function. The action value function, also referred as the Q-value function, represents the sum of future rewards expected for taking a particular action $a$ in a particular state $s$ of the environment, and is denoted by Q(s,a). On the other hand, the state value function represents the sum of future rewards expected from a particular state of the actor’s environment. It is often denoted by V(s). If an actor always chooses only one action in a given state, then its action value function would
be equal to the state value function. Otherwise, the state value function would refer to
the average of action value functions weighted by the probability of taking each action in a
given state of the environment. 

A general architecture of reinforcement learning is presented in Fig.~\ref{fig:rl}, where an agent interacts with an environment. This architecture has two main properties: learning and playing~\cite{sethyareinforcement}. In the continuous training phase of the model, initially the R-learner does not know which action will maximize the gain, as in most other machine learning. It has to discover which actions are most profitable by applying them. So it performs random actions from a state and observe the incurred reward (or penalty) using the feedback of that action from the environment. These rewards (or penalties) are the Q-values for particular state action pairs, which define the policy. The RL agent, while playing, uses this policy to predict which action is best in a given state and learns through the feedback from the environment.



\begin{figure}[t]
    \centering
    \includegraphics[width=0.6\columnwidth]{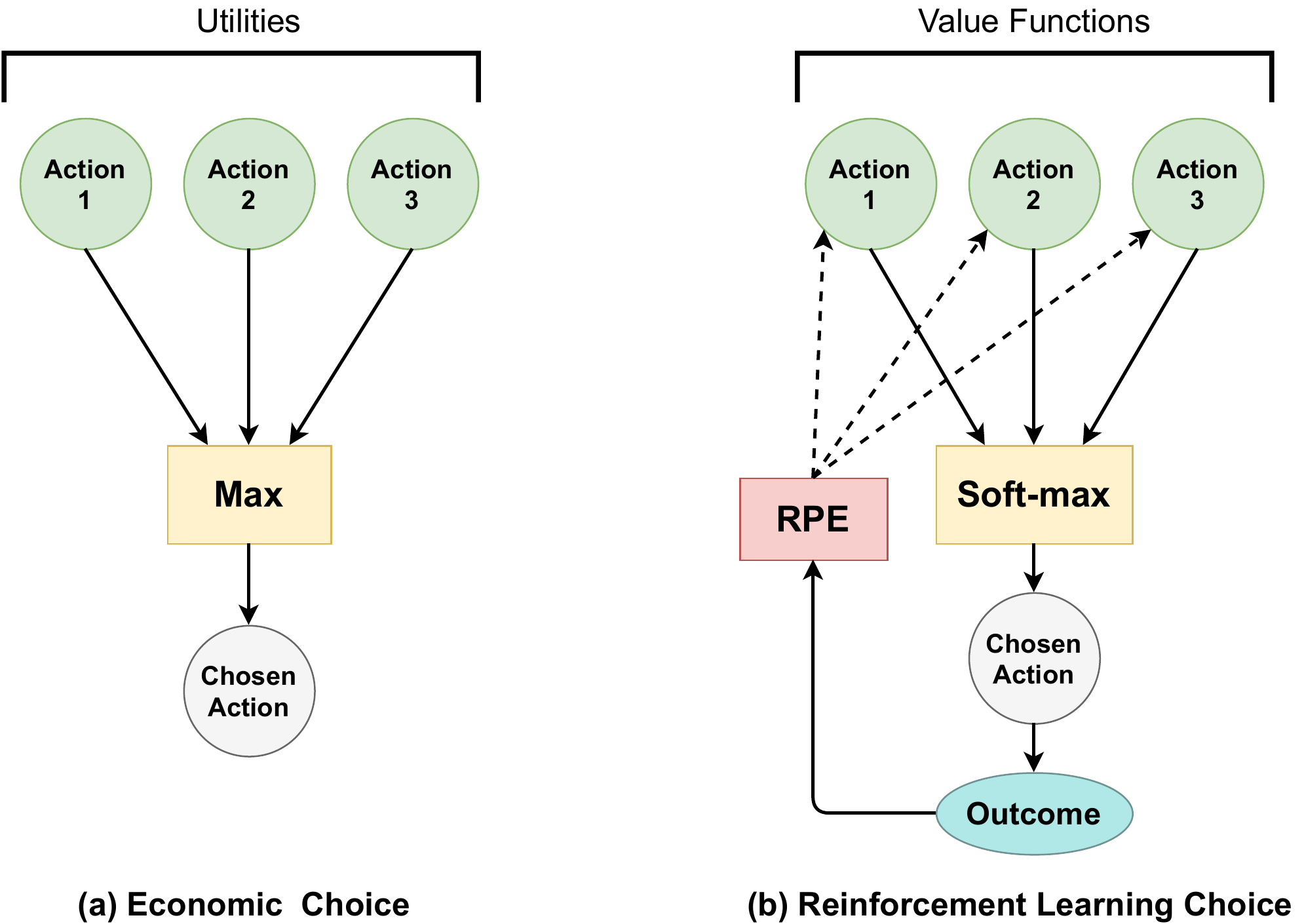}
    \vspace{-6pt}
    \caption{ Economic and reinforcement learning theories of decision making}
    \label{fig:eco_vs_rl}
    \vspace{-12pt}
\end{figure}

\subsection{Decision Making: Economic vs Reinforcement Learning}

Decision making refers to the mechanism that is used by an actor to choose its actions.
Economic theories of decision making attach numerical metrics to the alternative
actions. In this way, the choices for specific actions can be better understood as actions with maximum value will be selected more often among all possible actions. These hypothetical quantities are called utilities and can be applied to any kind of behavior. Fig.~\ref{fig:eco_vs_rl}(a) presents the economical choice of decision making, where the actor will always, by definition, choose the behaviors that maximize the organism’s utility~\cite{lee2012neural}. These theories behave agnostically about how these utilities are determined. However, they are presumably constrained by individual experience and evolution~\cite{lee2012neural}. Similar to utilities in economic theories, value functions in RL theory refer to the estimates for the sum of future rewards. As represented in Fig.~\ref{fig:eco_vs_rl}(b), an actor chooses the actions probabilistically (i.e., softmax) on the basis of their value functions, which is updated on the basis of the outcome (reward or penalty) resulting from the action chosen by the actor using a reward prediction error (RPE). 

Although both of these two approaches have their own unique benefits, one assigns more weight to the immediate reward while the other does it for the future rewards. In a partially observable environment, if an agent concentrates on both the immediate and future rewards before taking any action, it might incur an increased total reward at the end. In this work, we consider two types of action performed by the agent to maximize the reward. The agent will optimize the trajectory by leveraging Q-learning algorithm based RL and will trade POIs with other agents depending on the reward gain versus resource invested by utilizing the economic decision making.

\section{Related Works}
\label{sec:related_works}
Trajectory-planning is one of the most frequently explored problems in UAV domain. It is an inaugural part UAV path-planning problem, for which, neumerous research proposals have been performed. This problem has been explored in terms of different
complexities and shapes~\cite{nilsson1969mobile, thompson1977navigation, lozano1979algorithm, tokuta1998extending}. Like in~\cite{jiao2010research, li2011coverage}, the regions to be explored by the UAVs are described by different sweep direction
for finding an optimal path. For minimizing the distance between sub-areas of the whole trajectory, Torres et al. explored the  back and forth pattern~\cite{torres2016coverage}. For convenience of the trajectory-planning, many research works reform the area map into simpler shapes. The authors in~\cite{balampanis2016area, balampanis2017area, balampanis2017coastal, balampanis2017spiral} divide the coverage area into the exact shape (triangle) by using approximate cellular decomposition
technique and hybrid decomposition technique. Similarly, a spiral-like pattern is presented by Acevedo et al. in ~\cite{acevedo2013cooperative} for performing path planning in complex coverage areas. 

 Originally problems were solved by framing the problem as a convex optimization problem and utilized analytic and numerical techniques \cite{ahmadzadeh2006multi}. UAVs had fixed wings, a minimum turning radius, and had to constantly move forward, with goals set as information gathered and camera coverage making the solution fairly general for many surveillance missions. While it was computationally expensive, a path need only be calculated once and so in effect this was  not a serious problem \cite{ahmadzadeh2006multi}.

Modern UAV design made the multiple UAV case economical. While this paper is primarily about surveillance problems, its necessary to remark on the similarities between surveillance, forest fire monitoring \cite{vinogradov2019tutorial} \cite{ghamry2017multiple}, 5G load reaction, and IoT data extraction \cite{zeng2019accessing} \cite{li2018uav} problems as all having problem descriptions related to point of interest discovery, monitoring, and path planning in dense environments with many restricted or intraverse-able sections. The inherently dynamic nature of these applications where, a UAV swarm need not only react to a changing environment but to how its individuals members act in that environment to maintain efficient resource usage. This is reflective of the current treatment of UAV swarms as a collection of autonomous vehicles. Swarm Intelligence Optimization (SIO) methods treated each UAV as a single agent in a swarm, initially moving randomly but placing down a virtual "pheromone" which would influence later agents follow the same path~\cite{gu2018multiple}. Other works using SIO included a rule system on top on SIO to prevent pathological swarm behaviour, and increase performance \cite{alfeo2018swarm}. Further, UAVs must communicate with a centralized authority for pheromone data representing a significant security risk. It does however have the important benefit of being able to explore and adapt to new targets being found.


Model-free reinforcement learning methods have become very popular in the field of path planning~\cite{yan2019towards} as it enables an agent to autonomously learn an optimal policy through trial-and-error interactions with its environment.Similar to ~\cite{hsu2020reinforcement} we used Q-learning to calculate collision free paths, but unlike it we also used Q-learning to generate the initial paths. Like in~\cite{yijing2017q}, the authors proposed an adaptive and random
exploration (ARE) approach that enables UAV navigation while
avoiding obstacles. Li et al. introduced a method for path planning that combines an improved version of the Q-learning algorithm with heuristic
searching rules for mobile robots in a dynamic environment~\cite{li2015dynamic}. In~\cite{tang2017cyclic}, a Q-learning based distinct derived learning method with cyclic error correction has been proved effective
for mobile robot navigation. The authors in~\cite{yan2018path}  improved the performance of Q-learning algorithm with an action selection strategy and a Q-function initialization method, which has been applied to UAV path planning in an antagonistic environment.

The trading of tasks or goals in multi-objective games is not a new concept. Like in~\cite{tesauro2002pricing}, the authors set up agents in an economic game including trading product or capital and receiving reward based on performance in the economic game. 
They utilized Q-learning as opposed to directly solving for the Nash Equilibrium in-order to generalize the method to partially observable and unstable state spaces more typical of real economic situations. In~\cite{schultink2008economic}, Schultink et al. utilize a hierarchy of agents at different levels of control, working together to complete a task.
This is similar to how our bidding system rewards agents choosing to move closer to their POI, and facilitates exchange of POIs. However our agents are more autonomous, meaning the bidding process cannot select from a set of agents, and allow that agent to act. Further our system is multi-agent, where different policies affect the environment simultaneously. The authors in~\cite{khan2020trusted} used an auction mechanism in addition to support vector machines to assign UAVs services, optimizing cellular coverage while balancing trust and profit potential. Our work differs in solution technique, and architecture by simultaneously solving for path and point assignment. An auction system based on profit potential is proposed by Ng et al. in~\cite{ng2020joint}, constrained by resource and current objectives to distribute uavs to cells to complete federated learning tasks. UAVs were allowed to form cooperative teams, choosing based on profit potential. Whereas, in our work, we consider all the UAVs working together as group of individual trading agents. ~\cite{duan2019novel} introduced an auctioning system for uav task assignment, taking into account resource constraints in environments with varying profit potential, risk, and information. A priority based point selection mechanism is used there before assigning them to the UAVs. On the other hand, we consider visiting each of the points at the earliest possible time, not assigning low priority to any of them.

\section{Proposed \name{} Framework}
\label{sec:proposed_framework}

\begin{figure}[t]
    \centering
    \includegraphics[width=.6\columnwidth]{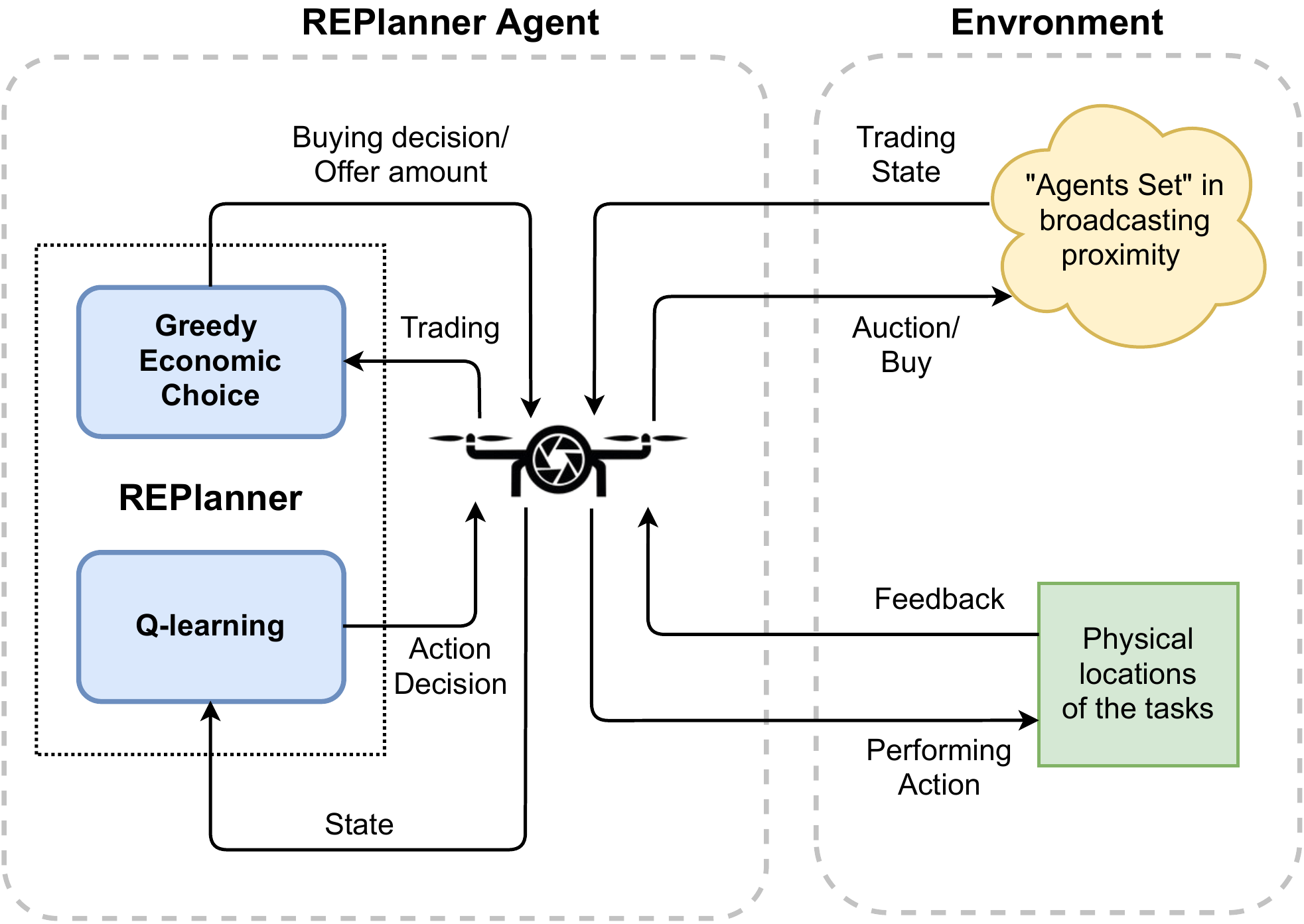}
    \vspace{-5pt}
    \caption{Proposed \name{} framework}
    \label{fig:frame}
    \vspace{-9pt}
\end{figure}

The basic framework of our proposed mechanism \name{} is presented in Fig.~\ref{fig:frame}, where the agent uses the 
Q-learning algorithm~\cite{watkins1992q} to take any kind of decision regarding it's movement and leverages greedy economic approach for trading POIs. The main idea of Q-learning is the model-free learning strategy where, without requiring an explicit environment model, learning associations between  state and action pairs and changes to the environment. The agent can take best actions under the given the observed, potentially noisy and information incomplete state. It does so by querying a table built from the result of these action and state pairs, and the consequent reward. Operating independent of this table is an auction system which modifies the environment in tandem with the Q-learning algorithm. In effect our agent acts upon two different perspectives of the environment at the same time. The first perspective of the environment consists of the physical locations that an agent has to travel within the time limit to gain it's rewards. The second perspective contains the offers and bids of all the agents in the environment. So effectively the agent, given that it is in two different but correlated states, must take the two most optimal actions in these two perspectives of the environment.

The agent uses the Q-learning module when it tries to find the optimal trajectory for it's visit to the points it's goal set. The Q-learning module takes state information of the agent as well as the partial information of the environment and calculates the trajectory incurring highest amount of reward. The agent takes action according to the output of the Q-learning module. On the other hand, an agent checks the feasibility of the points in it's goal set using the greedy economic choice (GEC) module to decide on which points to visit and which to sell off. The agents broadcast an auction for each of their infeasible points to the other agents that are within the broadcasting proximity. The other agents, listening to this broadcast, use their own GEC module to calculate the feasibility for the point of interest and estimates an offer price for that point. They broadcast the offer price to the selling agent and after finding the best offer, the selling agent transfers the point to the buying agent's goal set. An agent observes the environment and provides current position and trading state to the Q-learning module and GEC module respectively.


\section{Technical Details}
\label{sec:technical_details}

First we discuss the necessary formal models underlying the proposed \name{} and later, we discuss the model in details.


\subsection{Environment Model}

A UAV swarm carrying out a surveillance mission adds a number of non-trivial requirements to a swarm path planning problem. Targets can dynamically change their position, potentially in reaction to being targeted, and there may be targets with unknown positions. Further a swarm must dynamically assign targets while maintaining surveillance of sensitive points. 



We model the environment as two simultaneous partially-observable Markov games. Markov games are a generalization of a stochastic decision process to multiple agents. A partially observable game hides a subset of the state from the players, usually other player's states. Formally it consists of a tuple $(S_i,A_i,O_i,Z_i,T_i,R,b_i^{(0)},\gamma)$
\begin{itemize}
    \item $S_i$ is the set of states for player $i$.
    \item $A_i$ is the set of actions for player $i$.
    \item $O_i$ is the set of observations for player $i$.
    \item $Z_i(o_i| \textbf{s}_{-i}, \textbf{a}_{-i})$ is the probability to generate an observation $o_i$ given that the vector of other players $\textbf{-i}$ played the vector of actions $\textbf{a}_{\textbf{-i}}$ when in the vector of states $\textbf{S}_{\textbf{-i}}$.
    \item $T_i(s_i^{'}|s_i, a_i, o_i \textbf{o}_{\textbf{-i}} )$ is the probability of transition from $s_i$ to $s_i^{'}$ given the action $a_i$ and observations $o_i$ and $\textbf{o}_{\textbf{-i}}$.
    \item $R_i(\textbf{S}\textbf{A})$ is the reward of player $i$ when the joint actions of all players $\textbf{A}$ are taken in the joint state $\textbf{S}$.
    \item $b_i^{(0)} \in \delta( \textbf{S}_{\textbf{-i}})$ is the initial belief of player i over the other players states. 
    \item $\gamma \in [0,1)$ is the discount factor of future rewards.
\end{itemize}
This formalization is able to encapsulate the dynamics of a UAV swarm carrying out a surveillance mission as agents change routes, targets change position, areas lack of information, and agents cooperate and compete for reward. There are two primary problems, assignment of POIs to agents and path planning for each agent to each of its assigned POIs. 
\subsection{The Path Planning Game model}
In a path planning game, agents must make a sequence of movement decisions in-order to navigate from POI to POI attempting to gather data before all other agents. The agents must avoid no-fly zones and agent-agent collisions. High priority points must be pursued by several agents to maintain up to date data. The agents are deployed onto a $W \times L \in \mathbb{N}^2$ grid-world with width $W$, and length $L$. POIs distributed randomly throughout with positions given by $POI :=\{(x_1,y_1),(x_2,y_2),...,(x_j,y_j)\}$ where, $(x,y) \in \mathbb{N}^2$ and $j$ is the number of POIs. Similarly no-fly zones are given by $\{(x_1,y_1),(x_2,y_2),...,(x_m,y_m)\}$ where $m$ is the number of off limits spaces with $(x,y) \in \mathbb{N}^2$. Agents positions are pairs 
$(x_i, y_i)$ where $0 < i < n$ and $n$ is the number of agents. A Player receives a positive reward inversely proportional to the time elapsed during the game when their coordinates match that of some element of $POI$, $P_l \in POI$. The POI is then removed from all agents goal sets and can no agent can receive  reward from traveling to it. 
Formally, a partially observable stochastic path planning game is defined as the following
\begin{itemize}
    \item $S_i:=(p_i,s_i)$ where $p_i\subseteq POI$ are the points the agent must visit and $s_i \in\mathbb{N}^2 $ is the agents position.
    \item $A_i:=m(s_i)$ if $s_{i+1} \notin N$ where $m$ adds or subtracts one from the elements of $s_i$ moving the agent to another square and $N$ is the set of girds composing a no fly zone.
    \item $O_i$ consists of the agent's state, $POI$ and the completeness status of each element of $POI$.
    \item $Z_i$ is defined implicitly based on the randomly initialized strategies of other players.
    \item $T_i$ is defined implicitly based on the actions of other players altering $POI$ and their own positions.
    \item $R_i$ is defined by a function which compares $p_i$ to $s_i$ and a function which compares $p_i$ to $\textbf{s}_{\textbf{-i}}$, whose weights are controlled by parameters $\alpha, \beta \in \mathbb{R}$ respectively. 
    \item $b_i^{(0)}$ is initialized randomly.
    \item $\gamma$ is set to 0.95.
\end{itemize}
The gridworld is initialized with $POI$, agent position, $p_i$, no-fly zones set randomly. It is represented by an image, in which different colors represent clear spaces, no-fly-zones, POIs, and UAVs. Agents are then allowed to move about in the grid world until all POIs are visited or a set time limit is reached. 
\begin{figure*}[t]
\centering
\includegraphics[width=.95\textwidth]{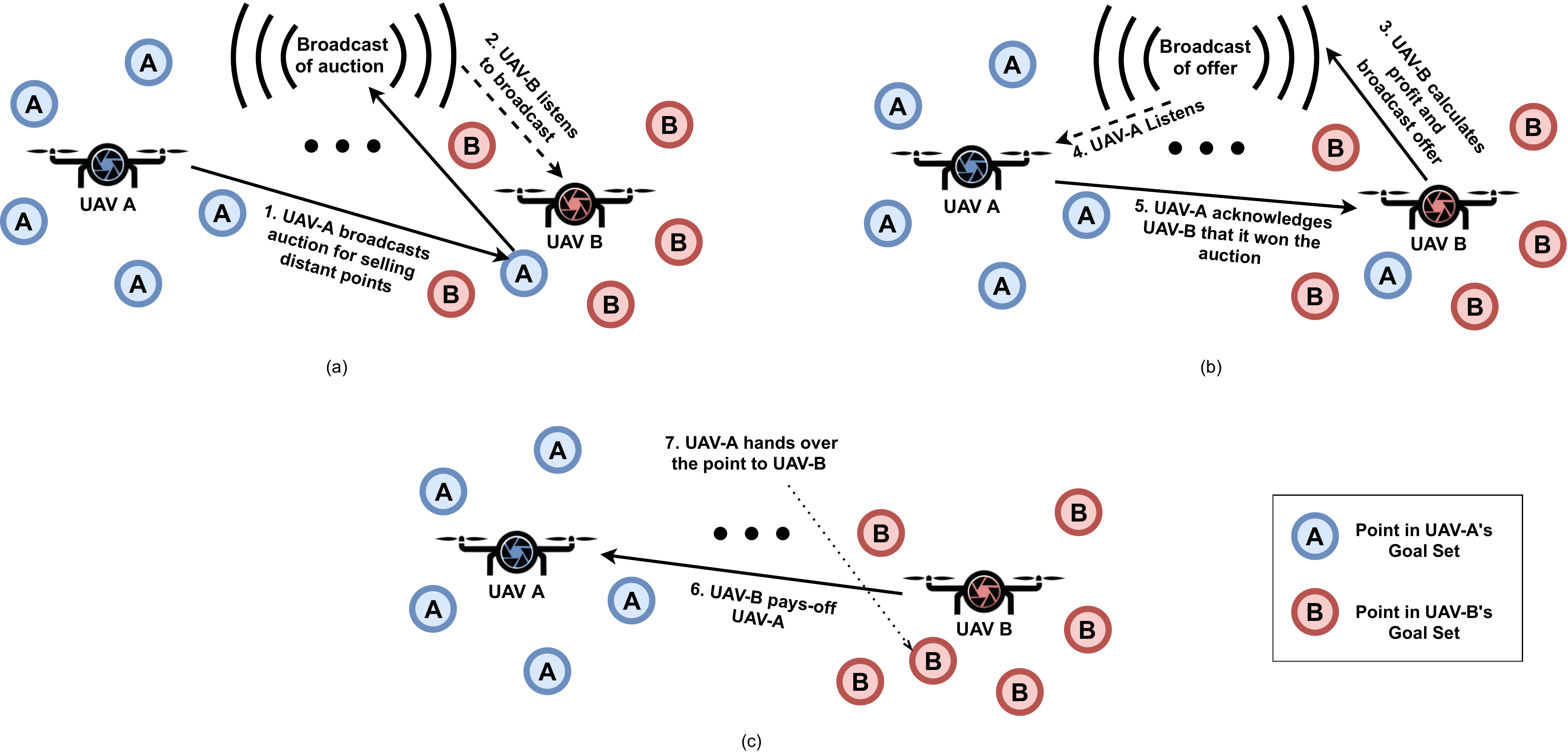}
\caption{Auction architecture : a) UAV-A broadcasts the auction for the distant point in its goal set while other UAVs listen to it and b) UAV-B determines its profit with respect to the expense-reward associated with the point of interest, and broadcasts its offer. Later, if it wins the auction, UAV-A acknowledges its victory. Finally c) UAV-B pays-off UAV-A to buy that point and accordingly, the point is handed over.}
\label{fig:architecture_n}
\vspace{-6pt}
\end{figure*}
\subsection{The Economic Game}
Simple models, such as the laws of supply and demand, function to describe the aggregate behaviour of self-interested agents cooperating and competing. We attempt to emulate this situation as a stochastic game, in order to exploit this optimization process. In an economic game, agents trade objects using a virtual currency, attempting to accumulate the greatest amount of privately valuable objects and capital. Each round agents engage in an auction, putting up an object currently in their possession and bidding a private amount for another object at auction. Fig.~\ref{fig:architecture_n} represents the auction architecture followed by the UAV agents. This models a set of simple agents attempting to most effectively distribute a set of objects according to each agent's valuation of said objects. Here we use an auction based on the resources an agents is likely to spend pursuing a point. This allows an agent to honestly bid it's valuation of the object while still being able to profit by selling the object at future auction for its perceived value. As agents have different valuations of each object, this allows for a stable distribution of objects to emerge over successive auctions. Accordingly there is a finite set of objects $O := \{o_j\}$ and a corresponding set $W_i:=\{o_k\in0|$ $o_k$ is owned by some agent $k \}$ , and exchange function $E$ is defined as: 
\begin{align*}
    E_i(o,P,A):= argmax_P(P,A) \xrightarrow[]{} W_a \cup \{o\} , W_i \setminus \{o\}
\end{align*}
where $o$ is the object for sale, $P$ is the set of offered prices, $A$ is the set of agents who offered prices. For example, the first price is offered by the first agent in $A$ and so on.
Formally an economic game is defined as the following
\begin{itemize}
    \item $S_i := (W_i, c_i)$ where $W_i$ is the list of owned objects as defined above, and $c\in \mathbb{R}^+$ represents the capital an agent has in trading.
    \item $A_i := (o_j,o_k,p)$ where $o_j \in O$ is an object selected for purchase, $o_k \in W_i $ is an object owned by the agent selected for sale, and $p \in \mathbb{R}^+$ is the price proposed for $o_j$.
    \item $O_i$ consists of the agents state and the objects for sale.
    \item $Z_i$ is based on the bids of other agents, and the objects in their possession which they decide to sell.
    \item $T_i$ is based on the exchange function $E_i$ defined above, assigning objects to the winners of bids, and exchanging capital between the seller and buyer of the object. 
    \item $R_i$ is proportional to the size of $W_i$ and $c_i$.
    \item  $b_i^{(0)}$ is initialized randomly.
    \item $\gamma$ is set to 0.95.
\end{itemize}
\subsection{Simultaneous games}
Both the economic and the path planning games are simultaneously played out during a UAV surveillance mission. In the path planning game, UAVs attempt to travel to POIs in the shortest time possible and before other agents. In this game, agents have no method to agree on POI responsibility and so, they compete for lucrative points, at the risk of wasting time and resources. We introduce a data structure called a Contract and assign these to be the objects traded in an economic game. Each POI has a corresponding contract. A UAV is rewarded for travelling to the POIs that it has contracts for. A UAV is forced to sell points too far away to be reached with its current goal set. This helps avoid overloading single UAVs with too many POIs. UAVs, which notice auctions for POIs near their current set of contracts, are incentivised to bid for those contracts. Playing both games simultaneously allows us to solve both point distribution and path planning problems in a distributed manner. 

\subsection{Full Environment Model}
A high level view of the architecture is as follows: We create a number of agents, which at a bare minimum must be able to buy and sell some object from each other. Our reward function is focused around this object, and whether holding this object nets a positive or negative reward is dependent on the agent's state. The agent can potentially prevent a negative reward by selling its object, but it can only be bought if another agent is willing to buy and so the purchasing agent must perceive acquiring the object as advantageous. This process distributes these objects optimally for every agent according to how the reward function treats these objects. Further, each agent is learning its own strategy from its experiences with the environment and other agents. As these objects are in fact contracts assigned to POIs with time and completion incentives, they distribute the contracts amongst themselves according to what can be travelled to most quickly.
\begin{itemize}
    \item $S_i := (W_i, c_i, s_i)$ where $W_i$ is the list of owned Contracts as defined above, and $c\in \mathbb{N}^+$ represents the capital used to trade contracts and $s_i \in \mathbb{Z}^2$ is the UAV's position.
    \item $A_i := (o_j,o_k,p,d)$ where $o_j \in O$ is an contract selected for purchase from the list of auction broadcasts, $o_k \in W_i $ is an contract owned by the agent selected for sale, and $p \in \mathbb{R}^+$ is the price proposed for $o_j$ and $d$ is the direction of the adjacent grid in $\mathbb{N}^2$.
    \item $O_i$ consists of the agents state and the contract sale broadcasts.
    \item $Z_i$ is based on the bids of other agents, and the objects in their position which they decide to sell, which depends on the other agents' valuation of the Contract which in turn depends on their owned contracts and current position.
    \item $T_i$ is based on the exchange function $E_i$ defined above, assigning objects to the winners of bids, and exchanging capital between the seller and buyer of the object. 
    \item $R_i$ is proportional to the size of $c_i$, and $1/|W_i|$ i.e., the reciprocal of the distance from each contract, and contract rewards
    \item  $b_i^{(0)}$ is initialized randomly as contracted are initially distributed evenly across all agents.
    \item $\gamma$ is set to 0.95
\end{itemize}


\subsection{Optimization problem}


Time is used as a proxy to real world quantities such as battery life and distance travelled, as these metrics are necessarily increasing functions of time. Thus by minimizing mission time we also minimize distance travelled and battery used. We enforce a hard time limit for the mission duration, to prevent battery over-use. We assume all UAVs are able to communicate with one another at any given time during auction periods.  
They move to an adjacent grid from their current position if the grid is accessible, according to:

\[
    s_{i+1}= 
\begin{cases}
    M(P_i,U_i ) = U_i + d_i,& \text{if } s_{i+1} \notin N\\
    s_i,              & \text{otherwise}
\end{cases}
\]


Redundancy constraints are introduced in the form of multiple contracts corresponding to a single POI. This guarantees that as many UAVs as there are contracts for a particular POI will visit.

As small UAVs have limited battery capacity, the goal is to visit as many points of interest as possible. Our solution focuses on maximizing the number of POIs travelled to in the least amount of time subject to redundancy and travel constraints. That is to say  $sup\{(\sum_{Contract \in C^*}c-t)   -T\}$, where $W$ is the collection of all contracts, $c$ is the contract's completion status, $t$ is the contract's elapsed time, and $T$ is the total mission time. 
\begin{figure*}[t]
    \begin{center}
    \hspace{-10pt} 
     \subfigure[]
        {
        \label{fig:case_eco}
            \includegraphics[width=0.49\textwidth, keepaspectratio=true]{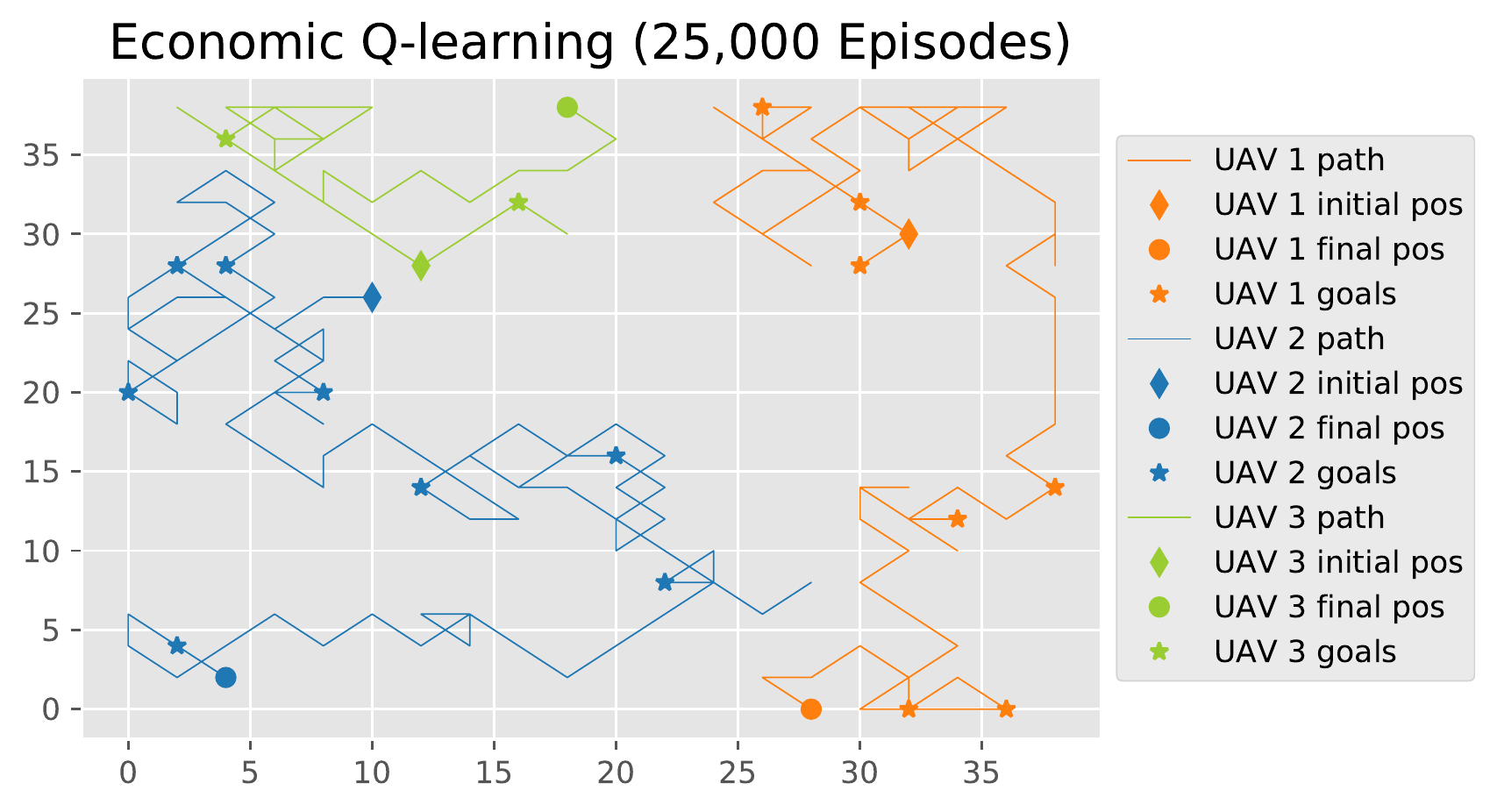}
        }     
        \hspace{-6pt} 
        \subfigure[]
        {
        \label{fig:case_reg}
            \includegraphics[width=0.49\textwidth, keepaspectratio=true]{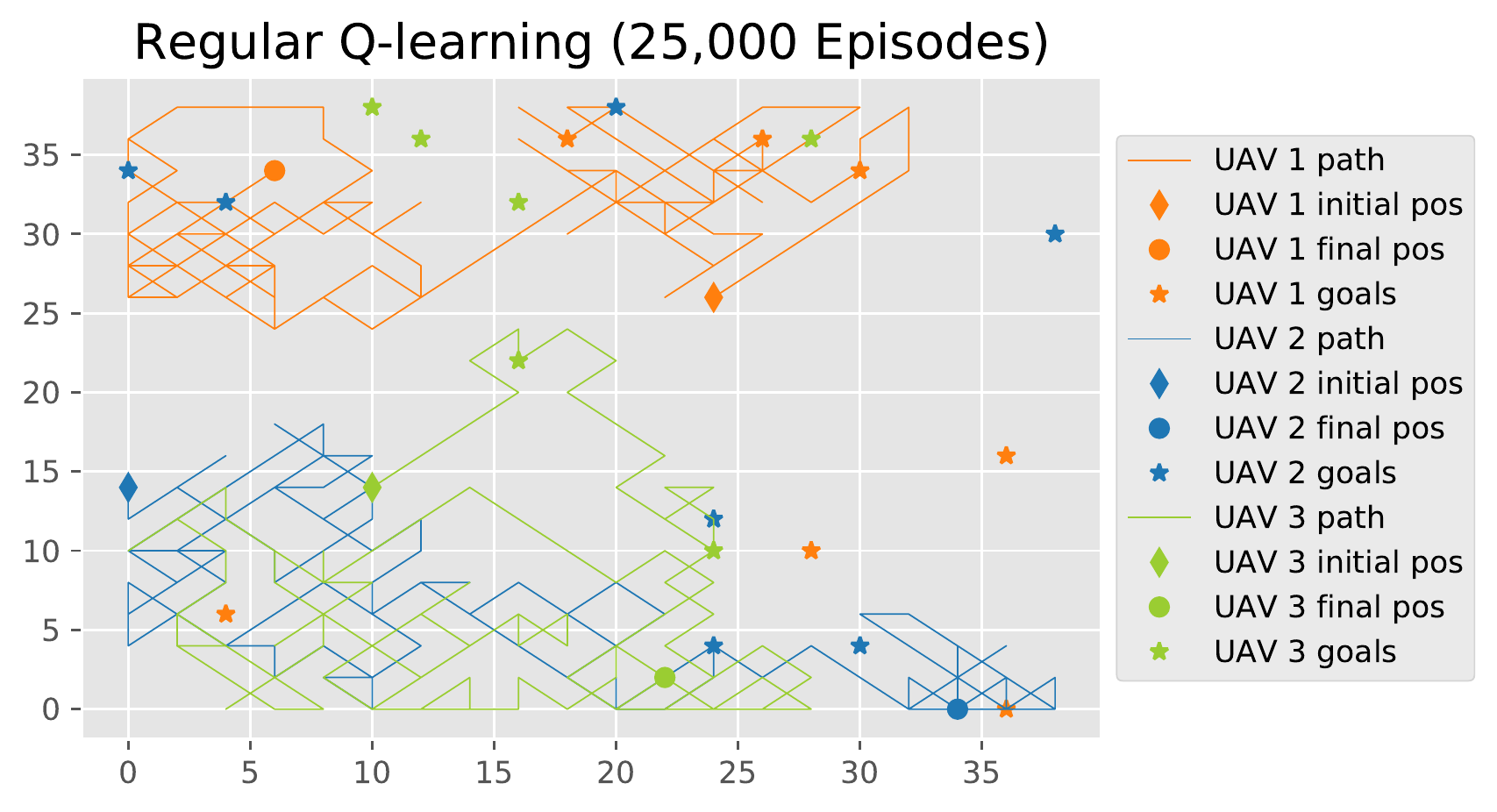}
        }
        \subfigure[]
        {
        \label{fig:case_reg2}
            \includegraphics[width=0.5\textwidth, keepaspectratio=true]{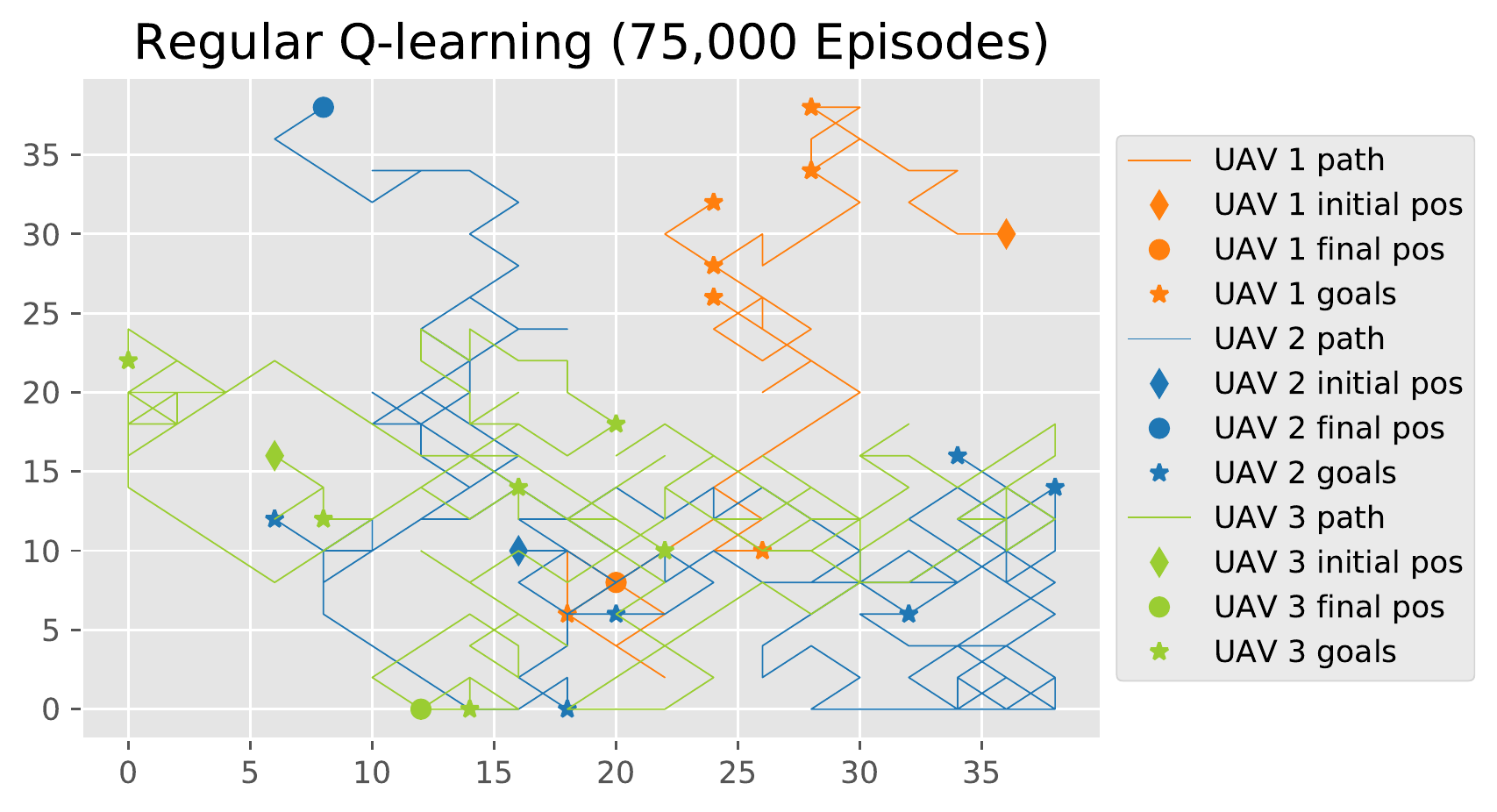}
        }
    \end{center}
    \vspace{-15pt}
    \caption{Case study of (a) Economic Q-learning after 25000 training episodes, (b) Regular Q-learning after 25000 training episodes, and (c) Regular Q-learning after 75000 training episodes with 3 UAV agents and 20 points to visit.}
    \label{fig:case_study}
    \vspace{-9pt}
\end{figure*}

\begin{table}[t]
\caption{List of Notations}
\vspace{-9pt}
\label{vartable}    
\centering
\begin{tabular}{|l|p{9.5cm}|}
\hline
Symbol & Definition \\
\hline
$A$ & Set of agents in the network \\
$a_i$ & The ith agents in the network \\
$R_{info}$ & Reward Information \\
$P_{info}$ & Price Information \\
$E$ & Environment data \\
$N_t$ & Number of training episodes \\
$O_{i}$ & Object list of ith agent, $O_{i}$ = [$o_{i1}$, $o_{i2}$,$o_{i3}$ ... ]\\
$O_{all}$ & List of all "object list"s, $O_{all}$ = [$O_{1}$, $O_{2}$, $O_{3}$ ... ] \\
$B_{i}$ & List of incoming bids of ith agent\\
$R(a)$ & Reward for completing action a\\
$R_i$ & Total reward of ith agent\\
\hline
\end{tabular}
\vspace{-9pt}
\end{table}

\begin{algorithm}[!t]
\SetAlgoLined
\scriptsize
\vspace{-2pt}
initialize $List\_{BroadCast}$ = $\Phi$;\\
\For{each episode in $N\_t$}{
    initialize $A$ = $init$($object$($R\_{info}$,$P\_{info}$),$E$);\\
    
    \While{$O\_{all}$ is not empty}{
        \For{each $a\_i$ in $A$}{
            \For{each $object$ in $O\_i$}{
                \If{$object$ is for not completed}{
                    $List\_{BroadCast}$.$append$($object$);\\
                }
            }
        }
        \For{each $a\_i$ in $A$}{
            \For{each $object$ in $List\_{BroadCast}$}{
                $B_i$.$append$($Distance$($a\_i,o\_i$), $object$ )
                
            }
        }
        \For{each $a_i$ in $A$}{
            
            choice, distance= $argmin_{distance}(B\_i)$;\\
            Update $O_i$ adding choice;\\
            Remove choice from $O_{seller}$;\\
            $R_{seller} +=10$;\\
            $R_i-=10$;\\

        }
        \For{each $a_i$ in $A$}{
            Use Q-table to choose $object$ from $O_i$ to take action in environment;\\
            Take action;\\
            \For{each $object$ in $O_i$}{
                \If{object is completed}{
                    Check $object$ for $R$;\\
                    \If{completed by self}{
                        Add to $R_i$;\\
                    }
                    Remove $object$ from $O_{all}$;\\
                }
                \Else{
                    \If{Collision or No fly zone detected}{
                    Subtract from $R_i$
                    }
                    continue;\\
                }
            }
        }
    }
}
\For{each $a_i$ in $A$}{
    update Q-table;
}
\normalsize
\caption{Training of UAV Trajectory Planning}
\label{algo}
\end{algorithm}
\subsection{Description of the Algorithm}
Finally, we introduce a novel distributed machine learning algorithm in order to optimally allocate targets. We note that market economies can be characterized as a distributed computation by market members for the appropriate distribution and production of items, and that these computations can be represented as repeated games as described above. Algorithm~\ref{algo} describes the bidding, moving and training process of the \name{} framework. Table~\ref{vartable} represents the list of notations used in this algorithm. 
Each UAV is controlled by a RL agent. Its state consists of known targets and completion status, its position organized into a map. Its actions consist of travelling to a point, selling and removing a point from it's memory, or buying and adding a point to its memory. When another agent reaches a point, it receives a large reward and the point is removed from all other agent's memories. Every other UAV is then unable to bid for, and gain reward by traveling to the POI. The process of buying and selling points is done via broadcasts, and does not require any centralized authority. In simulation, however, it is necessarily centralized.
Due to the the various levels of cooperation and competition between lower level agents each attempting to maximize their particular reward over time, the swarm as a whole prioritizes the correct amount of UAVs to each particular point, in a way that leads to the maximum potential for success of the mission in the shortest amount of time. Importantly, this aspect of the optimization process is not directly computed, but is an emergent phenomenon resulting from the collective interactions between agents and their bids, and thus requires no extra processing. When trained models are deployed on real UAVs they need only consider small messages from other UAVs consisting of price data, and an ID. In~\cite{chang2020decentralized}, it is shown that the aggregate behaviour of the agents, when in the appropriate market structure, is in line with the super agent's global goal.

\subsection{A Case Study}
We initialized an environment with 20 points, set at random locations in a 40x40 grid. UAVs’ positions are initialized within the same grid. Each agent is assigned contracts randomly. UAVs are allowed to move in the same grid. Fig.~\ref{fig:case_study} represents the two-dimensional space, where the UAVs trained with economic Q-learning (Fig.~\ref{fig:case_eco}) and regular Q-learning (Fig.~\ref{fig:case_reg}-~\ref{fig:case_reg2}) move around to explore the environment and visit the goal points. Each colored line is a UAV’s path throughout the 200 timesteps in the twenty-five thousand-th episode. Assuming each grid is 1 meter, and a velocity of 5 meters/second, each timestep is 1/5 of a second, creating an episode length of about 40 seconds. In the non-economic scenario, all the agents, without any trading of POIs, try to visit each of the goal locations, being non-consideration of resource and time constraints. In the process, they try to navigate through the whole map. From Fig.~\ref{fig:case_reg}, we can see a lot of overlapping paths and a lot of distant covered. However, none of the agents are able to complete the whole goal set assigned to them. However, after seventy five thousand episodes in Fig.~\ref{fig:case_reg2}, all the goal sets are successfully completed. Even then, there are a lot of overlapping paths, wasting a lot of resource and suffering latency in goal completion. On the other had, using the economic model in Fig.~\ref{fig:case_eco}, all the agent learnt to trade the initially assigned POIs and they display an excellent optimization of resource utilization. There are no overlapping paths and they seem to cover just part of the whole map, yet completed all the goals assigned.



\section{Evaluation Setup}
\label{sec:evaluation_setup}

\begin{table}[b]
\vspace{-15pt}
\centering
\caption{Q-learning Module Architecture}
\vspace{-9pt} 
\label{tab:q_architecture}
\begin{tabular}{|l|c|c|}
\hline
\multicolumn{1}{|c|}{\textbf{Parameter Name}} & \textbf{Model hyperparameters} \\ \hline
\textit{Model}                                & Q-learning                            \\ \hline
\textit{Exploration paeameter ($\epsilon$)}                               & 0.5                              \\ \hline
\textit{Epsilon decay}                 & 0.9999                     \\ \hline
\textit{Discount factor}                              & 0.95                           \\ \hline
\textit{Episodes (Each iteration)}        & 25,000                          \\ \hline
\textit{Steps (Each episode)}                            & 200                            \\ \hline
\textit{Learning rate}                        & 0.1                           \\ \hline
\end{tabular}
\vspace{-9pt} 
\end{table}

This section presents the experimental setup and necessary evaluation metrics to assess our proposed \name{} framework’s performance. We implement the framework considering a simple two-dimensional environment with time constraints for the UAV agents. The experiments are conducted on Dell Precision 7920 Tower workstation with Intel Xeon Silver 4110 CPU @3.0GHz, 64 GB memory, 4 GB NVIDIA Quadro P1000 GPU.

\subsection{Environment Objects and Actions}
The environment contains three types of objects: the UAV agents, the POIs to be visited and the no-fly-zones (NFZ) to be avoided. Each type of object is initialized in the environment at a unique position to ensure a POI doesn't end up at the same location as a NFZ. The agent can take 4 diagonal movement actions for visiting the POIs depending on the appropriate (maximum or minimum) q-value for that state-action pair. Also, for exploratory behavior, the agent can randomly take an action towards horizontal or diagonal direction. So a total of 8 actions can be performed for movement. Now, for the trading mechanism, an agent can take 2 types of actions: sell or buy a POI. The agent decides to sell a point if the monetary value of the resource requirement is more than the potential reward to be achieved. Later, an agent decides to buy a POI if the potential reward is more than the total of buying price and resource invested to visit that point.


\subsection{Architecture of the Model}
Our model begins with defining the various hyper-parameters mentioned in Table~\ref{tab:q_architecture}. For the very first episode, the q-table is initialized with random values. For a 50$\times$50 grid environment, the q-table has more than 96 million cells, each representing an action-value pair. Meaning more than 96 million q-values to be tuned throughout the learning process. For achieving good amount exploration, the $\epsilon$ value is set to 0.5 initially and is decayed throughout the process.

\begin{figure*}
    \begin{center}
    \hspace{-30pt} 
     \subfigure[]
        {
        \label{fig:time1}
            \includegraphics[width=0.30\textwidth, keepaspectratio=true]{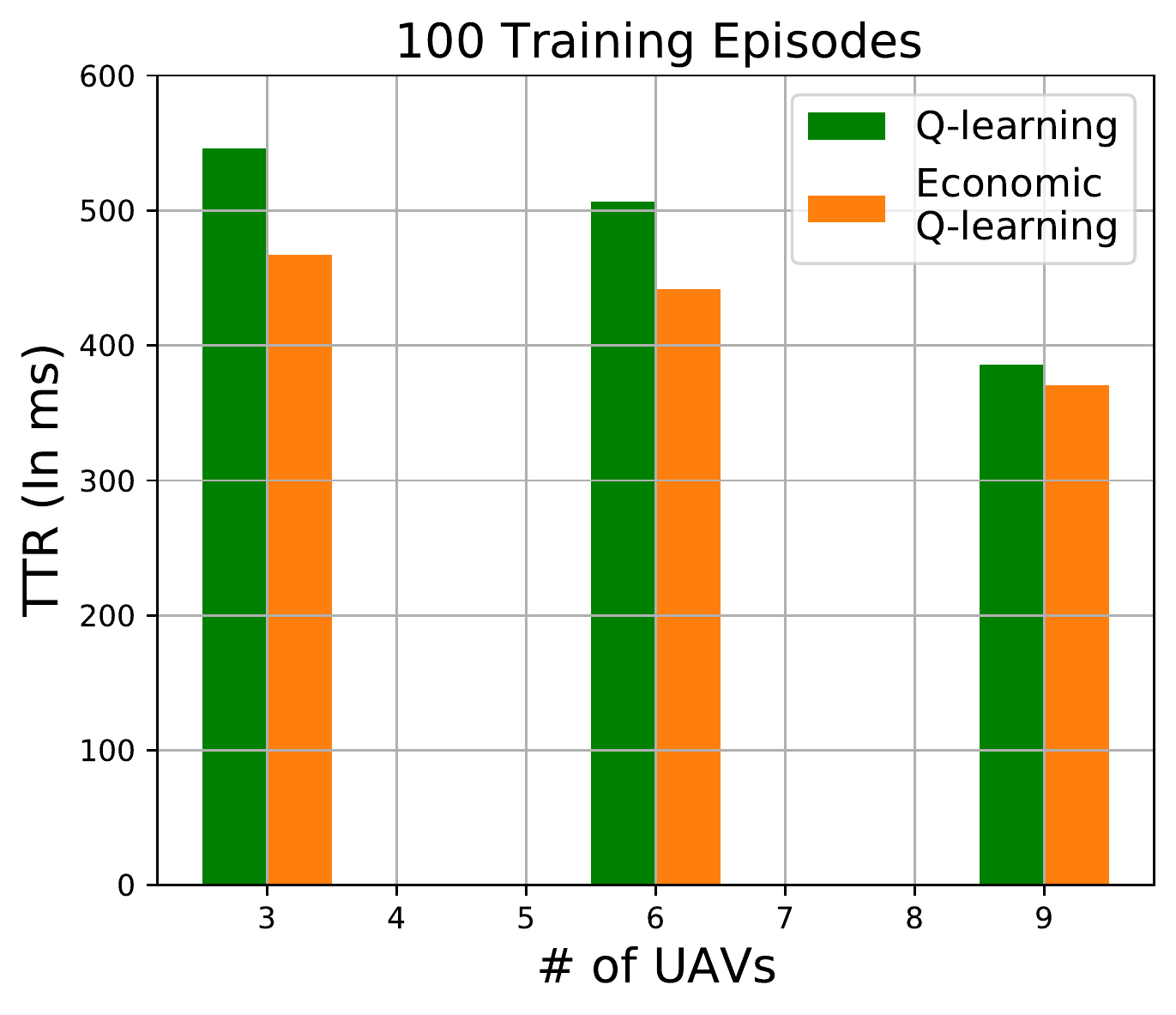}
        }     
        \hspace{-5pt}
        \subfigure[]
        {
        \label{fig:time2}
            \includegraphics[width=0.30\textwidth, keepaspectratio=true]{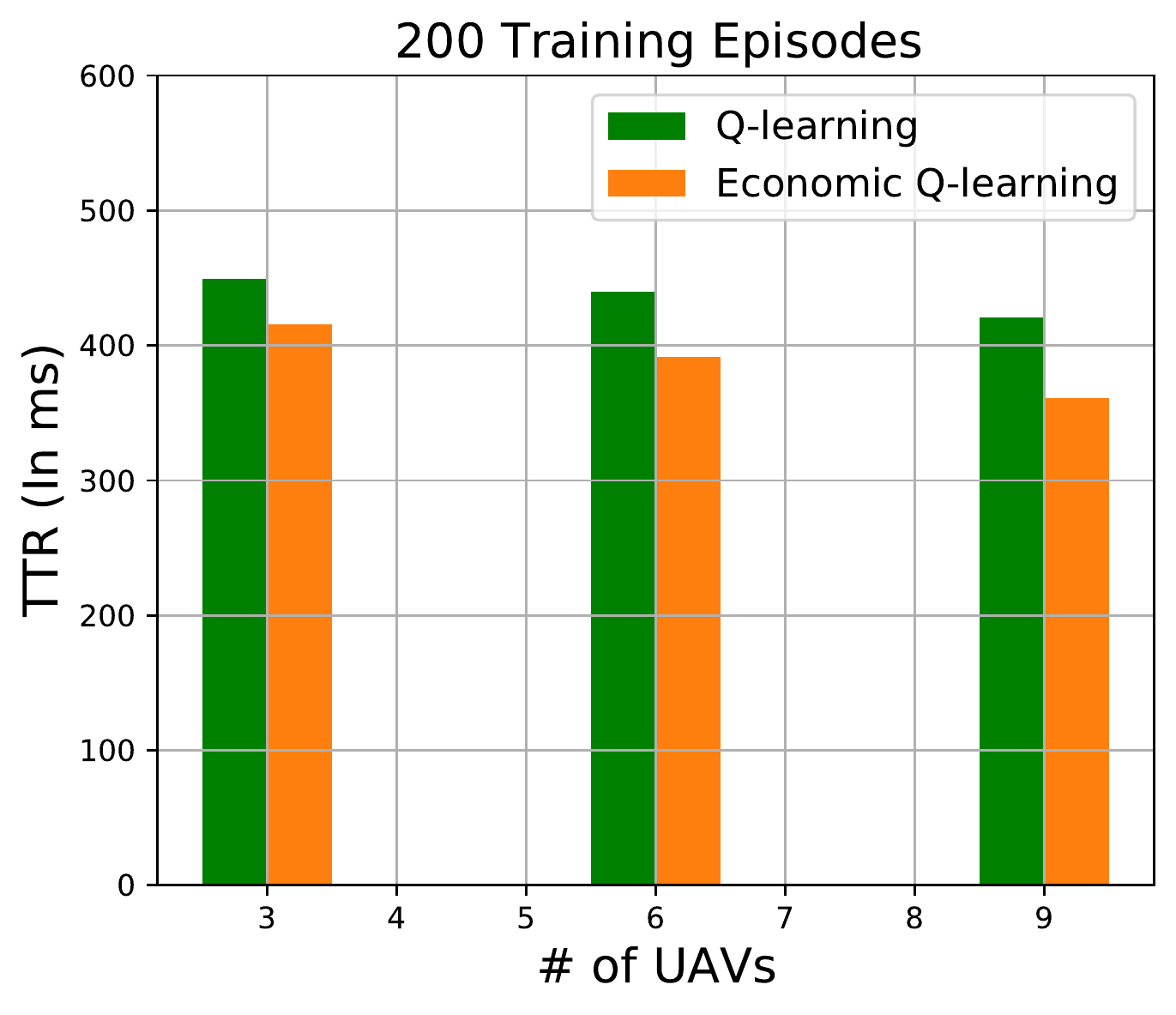}
        }        
        \hspace{-5pt} 
        \subfigure[]
        {
        \label{fig:time3}
            \includegraphics[width=0.30\textwidth, keepaspectratio=true]{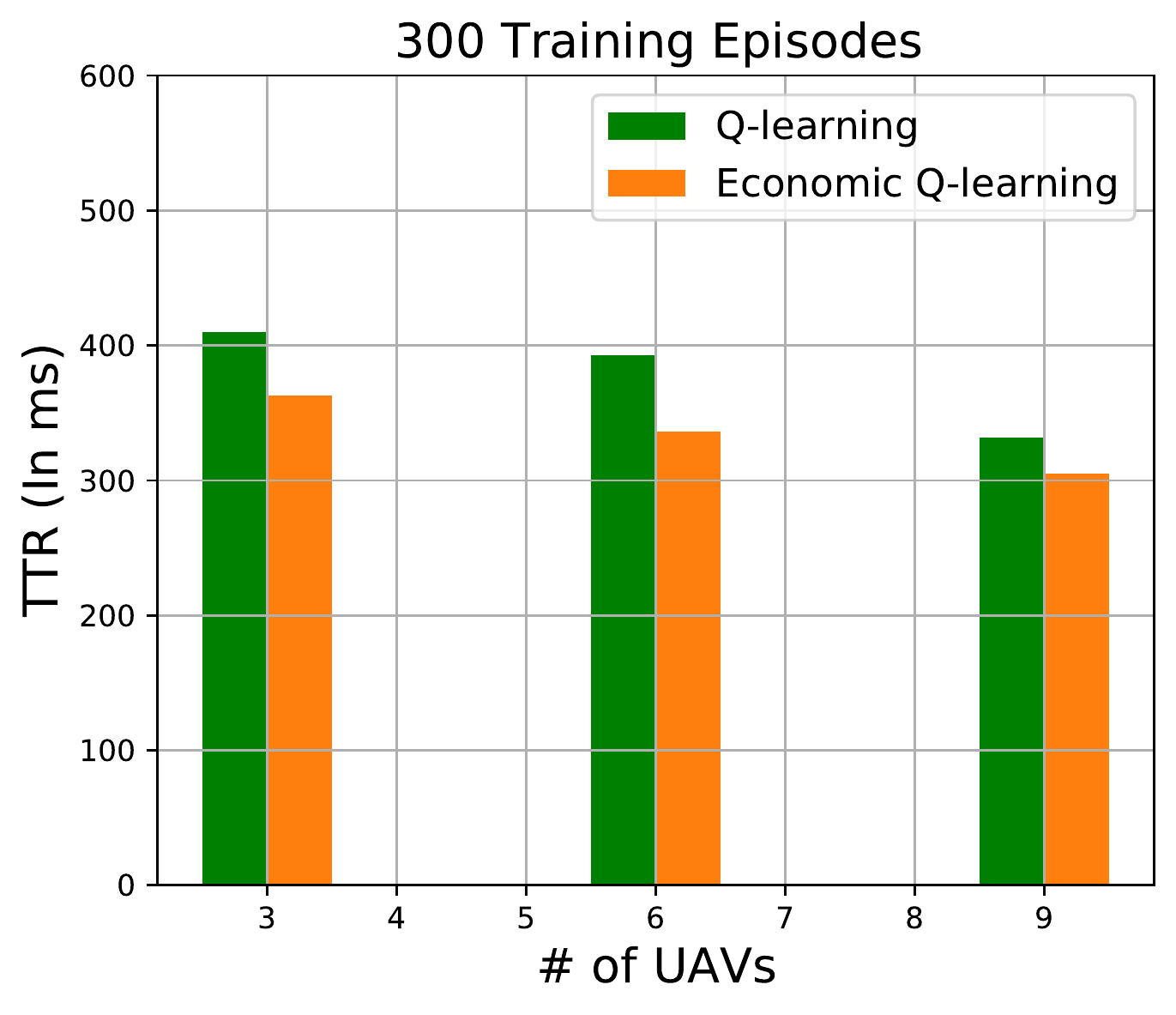}
        }        
    \end{center}
    \vspace{-15pt}
    \caption{Average time required to complete goal set after (a) 100 training episodes, (b) 200 training episodes, (c) 300 training episodes, for different UAV swarm with Q-learning and economic Q-learning visiting 20 POIs.}
    \label{fig:time}
    \vspace{-9pt}
\end{figure*}

\begin{figure*}
    \begin{center}
    \hspace{-10pt} 
     \subfigure[]
        {
        \label{fig:percent1}
            \includegraphics[width=0.32\textwidth, keepaspectratio=true]{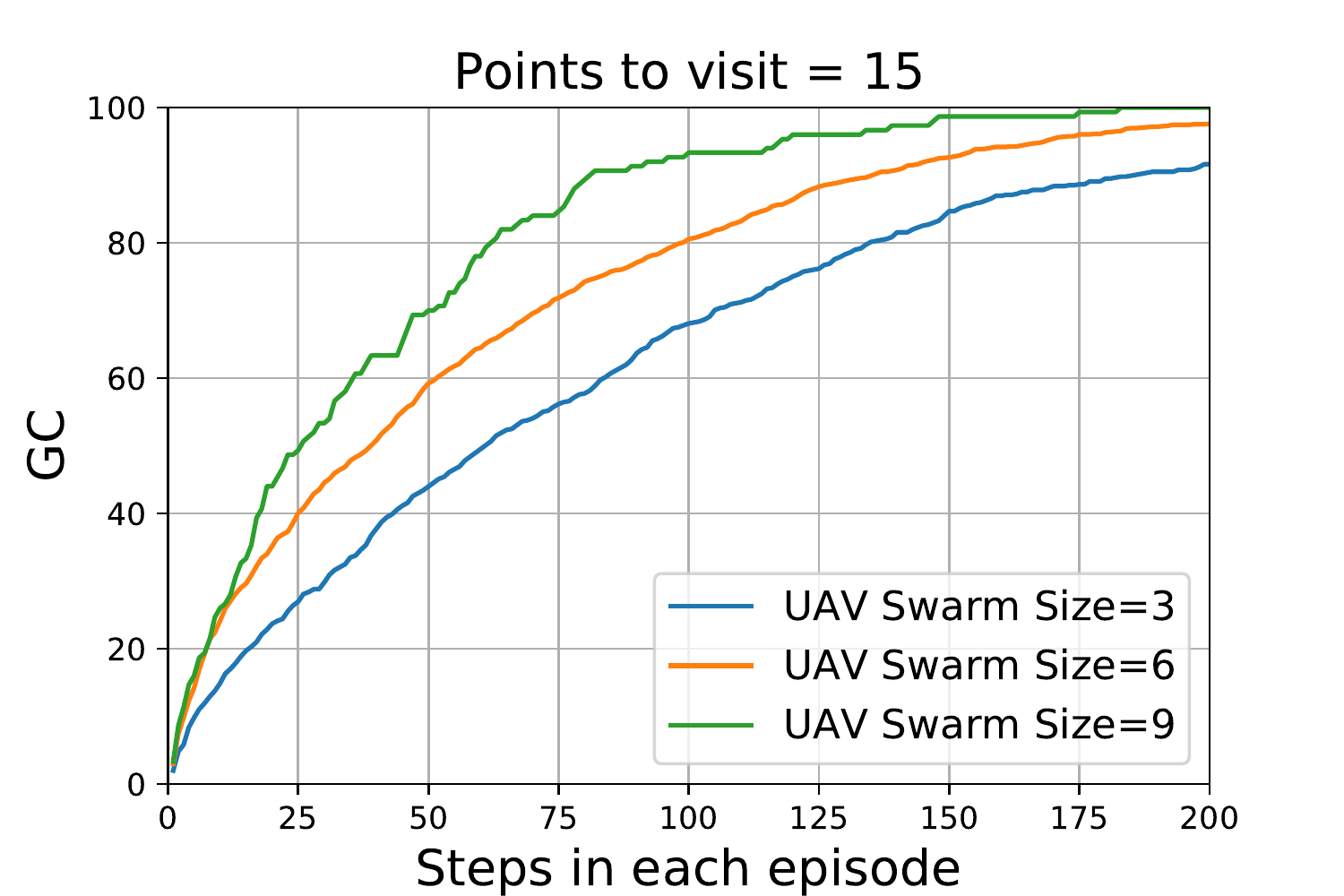}
        }     
        \hspace{-5pt} 
        \subfigure[]
        {
        \label{fig:percent2}
            \includegraphics[width=0.32\textwidth, keepaspectratio=true]{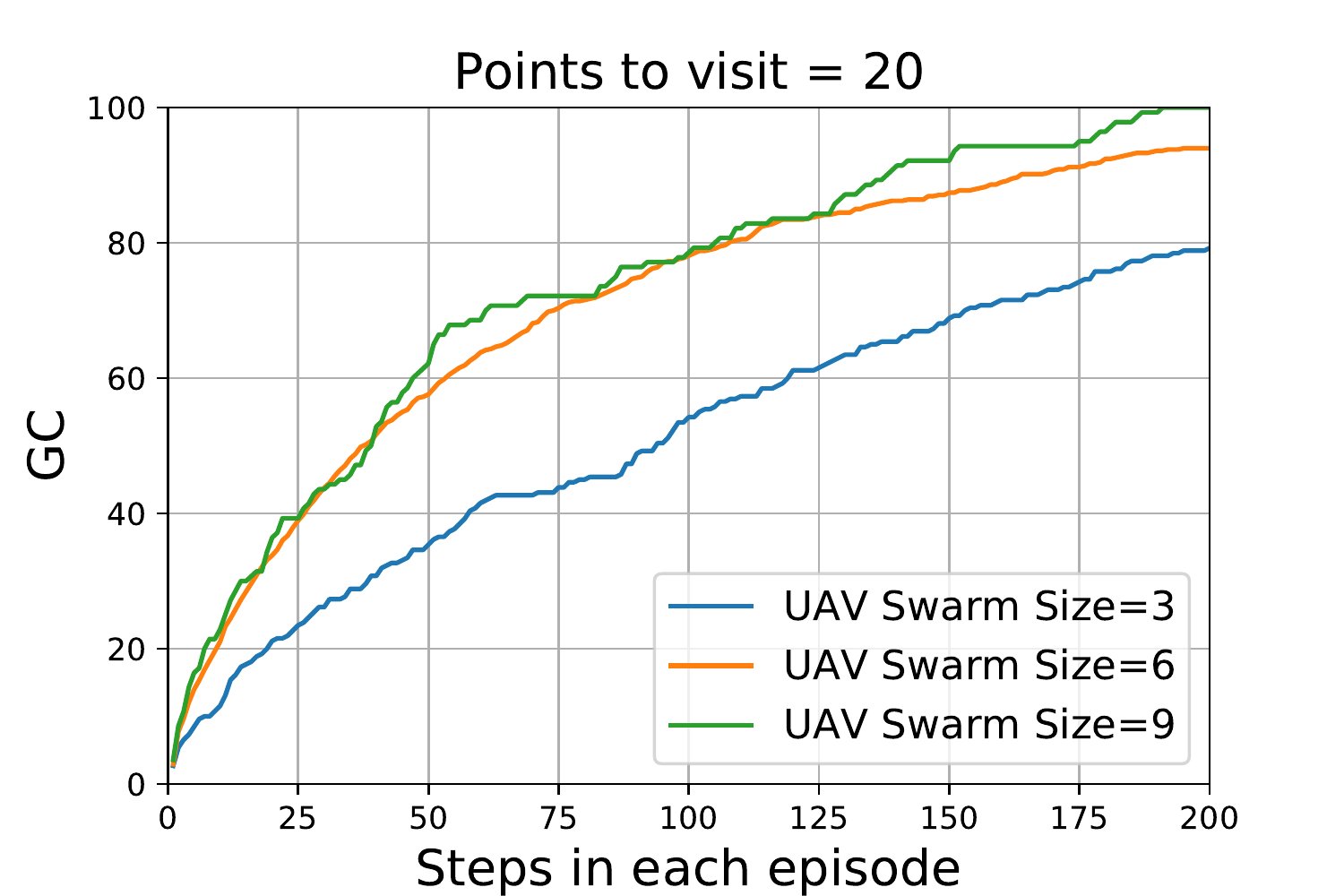}
        }        
        \hspace{-5pt} 
        \subfigure[]
        {
        \label{fig:percent3}
            \includegraphics[width=0.32\textwidth, keepaspectratio=true]{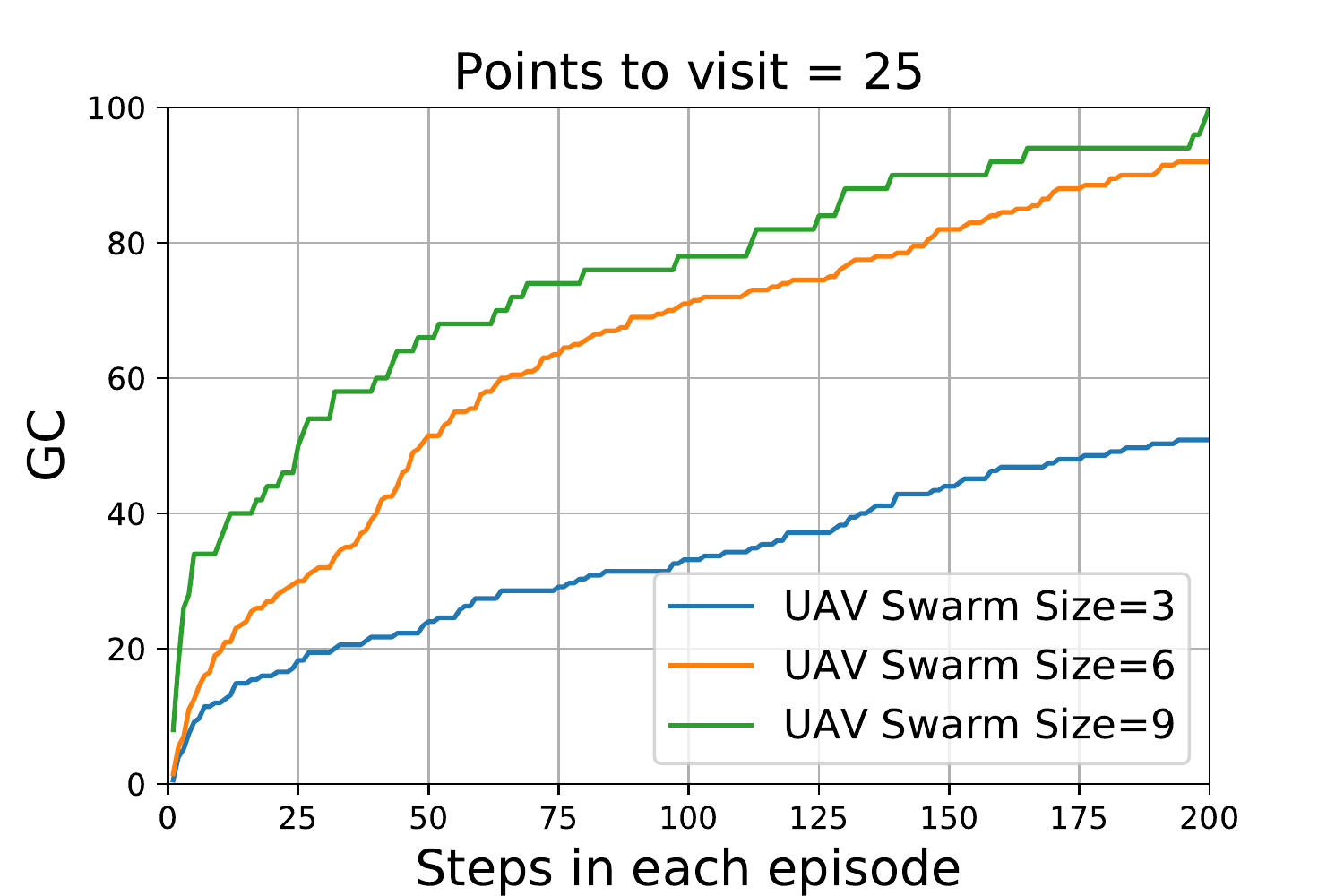}
        }        
    \end{center}
    \vspace{-15pt}
    \caption{Percentage of goal set completion with respect to number of steps taken for UAV swarm size of 3, 6 and 9 having (a) 15 points to visit, (b) 20 points to visit, (c) 25 points to visit, with \name{} framework after 1000 training episodes.}
    \label{fig:percent}
    \vspace{-9pt}
\end{figure*}


\subsection{Evaluation Metrics}
In this subsection, we define the metrics used to evaluate the \name{}’s performance.

\noindent \textbf{Total Time Required (TTR):} The efficiency of the protocol can be measured with respect to the total time required for all the agents to complete their goal-set. This is not the summation of total travel time of all the agents, rather the difference between the starting time and time when all the agents are done with their goal-set. Thus the TTR can be defined as the following:
\vspace{-5pt}
\begin{align*}
TTR = \textit{(final time-stamp)} - \textit{(initial time-stamp)}
\end{align*}

\noindent \textbf{Goal Completeness (GC):} GC is defined by the percentage of the goal set point that are already been visited or processed, depending on the context. Thus the GC can be defined as the following:
\vspace{-5pt}
\begin{align*}
GC = \frac{\textit{\# of points visited}}{\textit{\# of total points in the goal set}} \times 100
\end{align*}

\noindent \textbf{Distance Traveled (DT):} DT is defined by the summation of total distance the agents had to travel to complete their goal sets. Ideally, the smaller is value is, the more efficient the trajectory was. Thus the DT can be defined as the following:
\vspace{-5pt}
\begin{align*}
DT = {\sum_{i \in {\mathbb N}} { d}_{i}}
\end{align*}
Here, $d_i$ refers to distance that i-th agent had to travel to complete it's goal set and ${\mathbb N}$ refers to the total number of agents.


\noindent \textbf{Episode Average Reward (EAR):} Average of the rewards accumulated by all the UAV agents in a particular episode. The reward increases for a correct decision, whereas wrong decisions incur negative points. Therefore, the reward is a number indicating how good the model's prediction was on an unknown environment.  

\vspace{-5pt}
\begin{align*}
EAR = \frac{1}{{\mathbb N}} {\sum_{i \in {\mathbb N}} {r}_{i}}
\end{align*}
Here, $r_i$ refers to the incurred reward of the i-th agent at the end of a particular episode and ${\mathbb N}$ refers to the total number of agents participating in that episode.

\begin{figure*}
    \begin{center}
    \hspace{-10pt} 
     \subfigure[]
        {
        \label{fig:distance1}
            \includegraphics[width=0.32\textwidth, keepaspectratio=true]{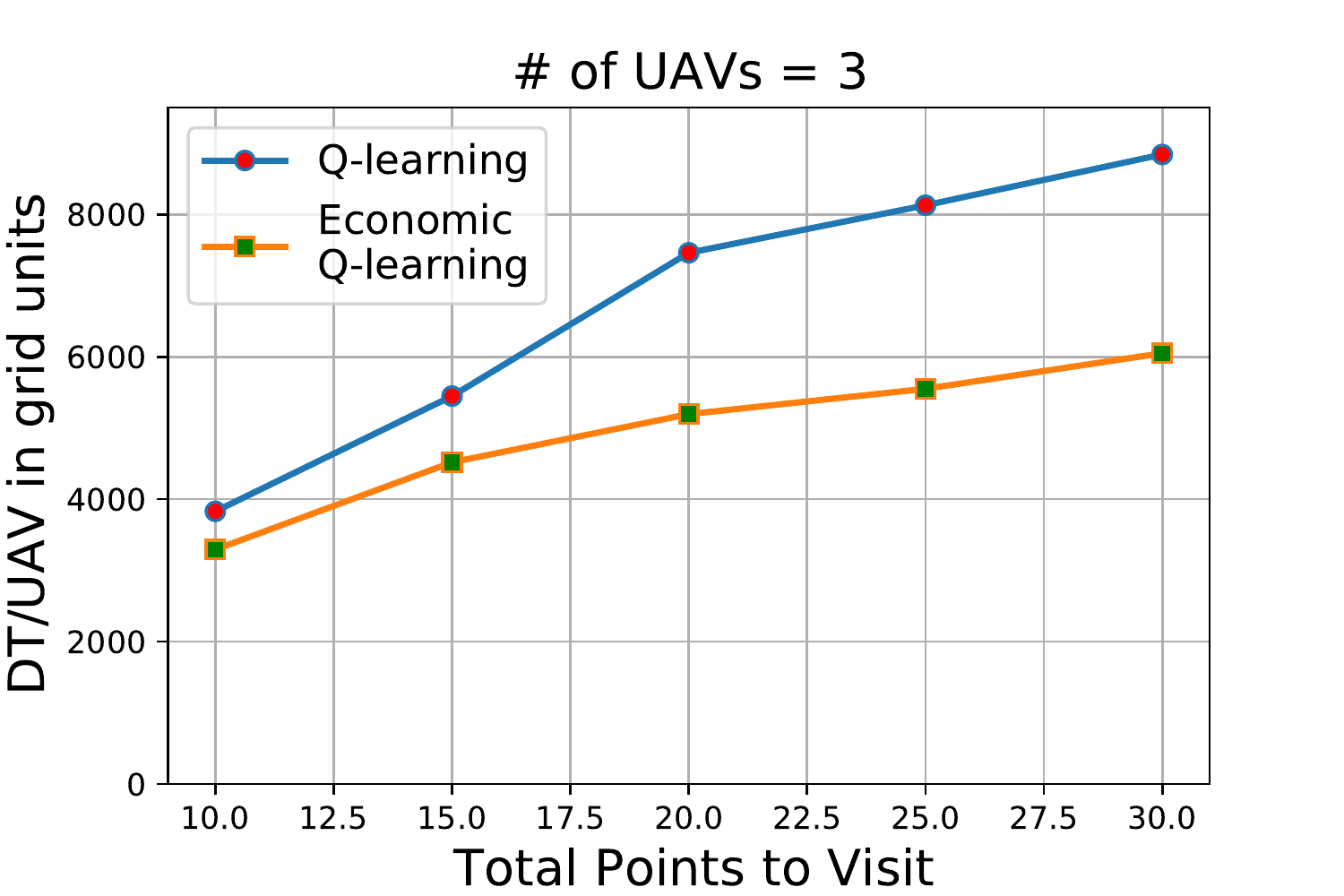}
        }     
        \hspace{-12pt} 
        \subfigure[]
        {
        \label{fig:distance2}
            \includegraphics[width=0.32\textwidth, keepaspectratio=true]{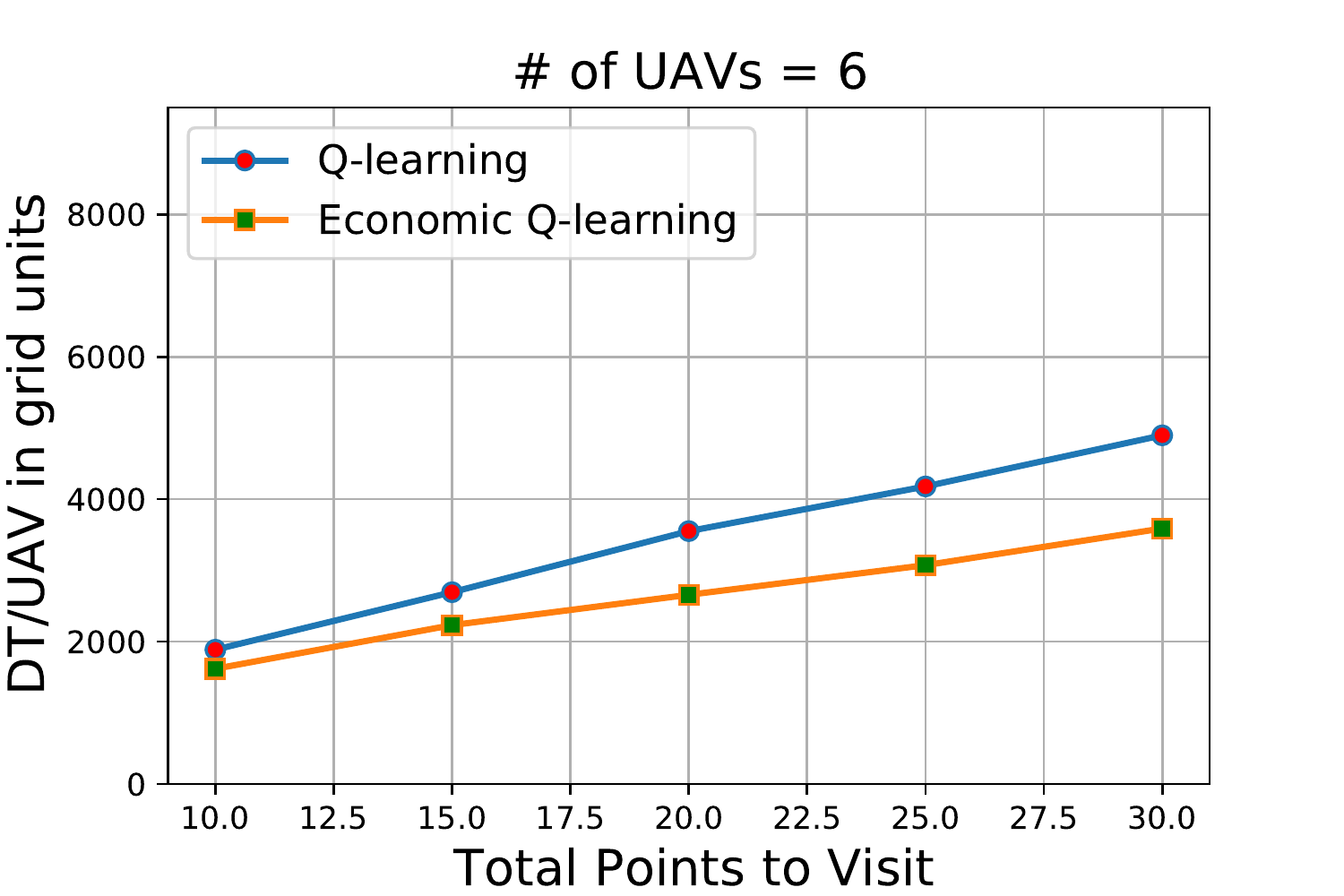}
        }        
        \hspace{-12pt} 
        \subfigure[]
        {
        \label{fig:distance3}
            \includegraphics[width=0.32\textwidth, keepaspectratio=true]{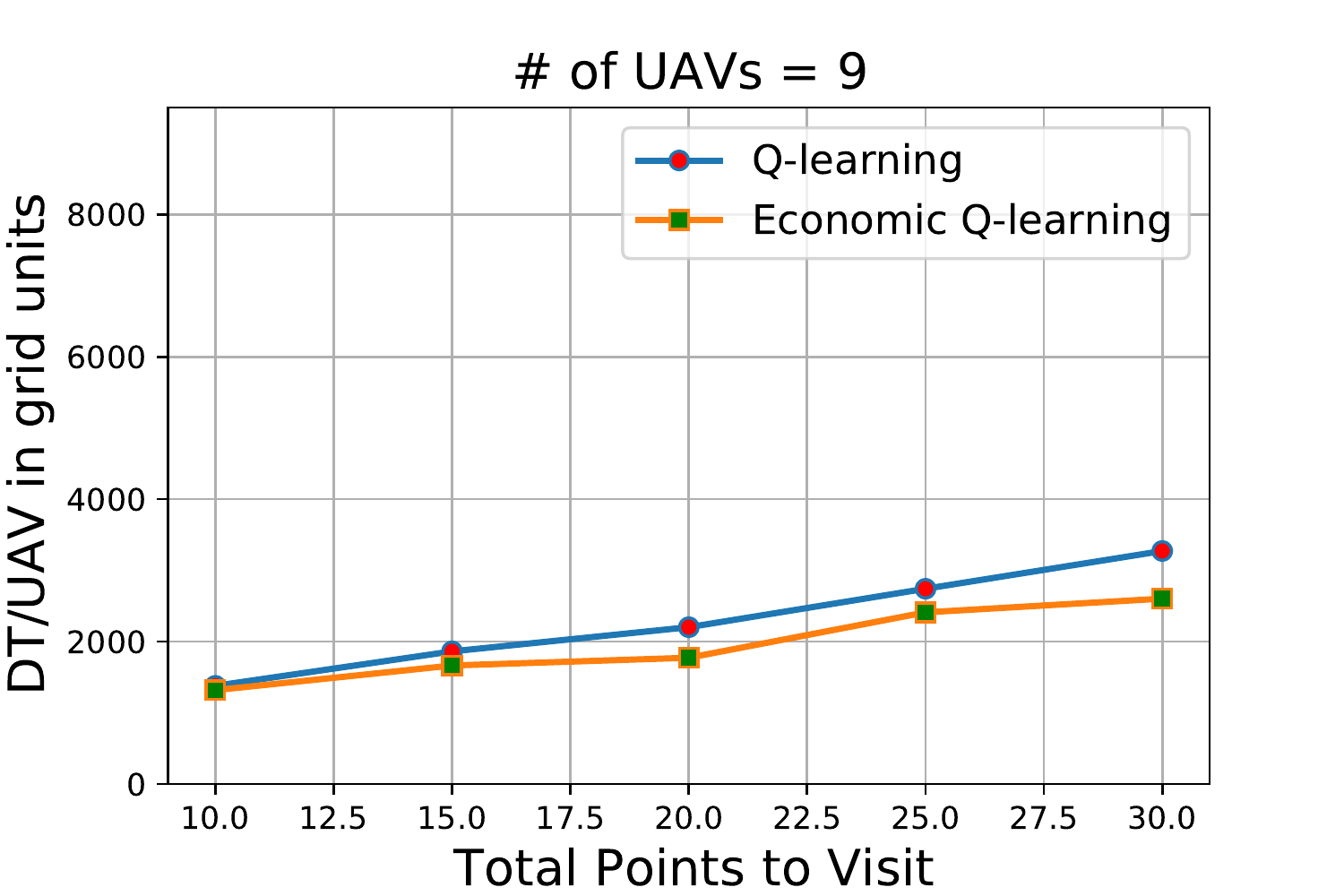}
        }        
    \end{center}
    \vspace{-15pt}
    \caption{Average distance required to be travelled to visit all the goal points with swarm size of (a) 3 UAV agents, (b) 6 UAV agents, (c) 9 UAV agents, trained with regular and economic Q-learning for 100 episodes.}
    \label{fig:distance}
    \vspace{-9pt}
\end{figure*}

\begin{figure*}
    \begin{center}
    \hspace{-10pt} 
     \subfigure[]
        {
        \label{fig:reward1}
            \includegraphics[width=0.32\textwidth, keepaspectratio=true]{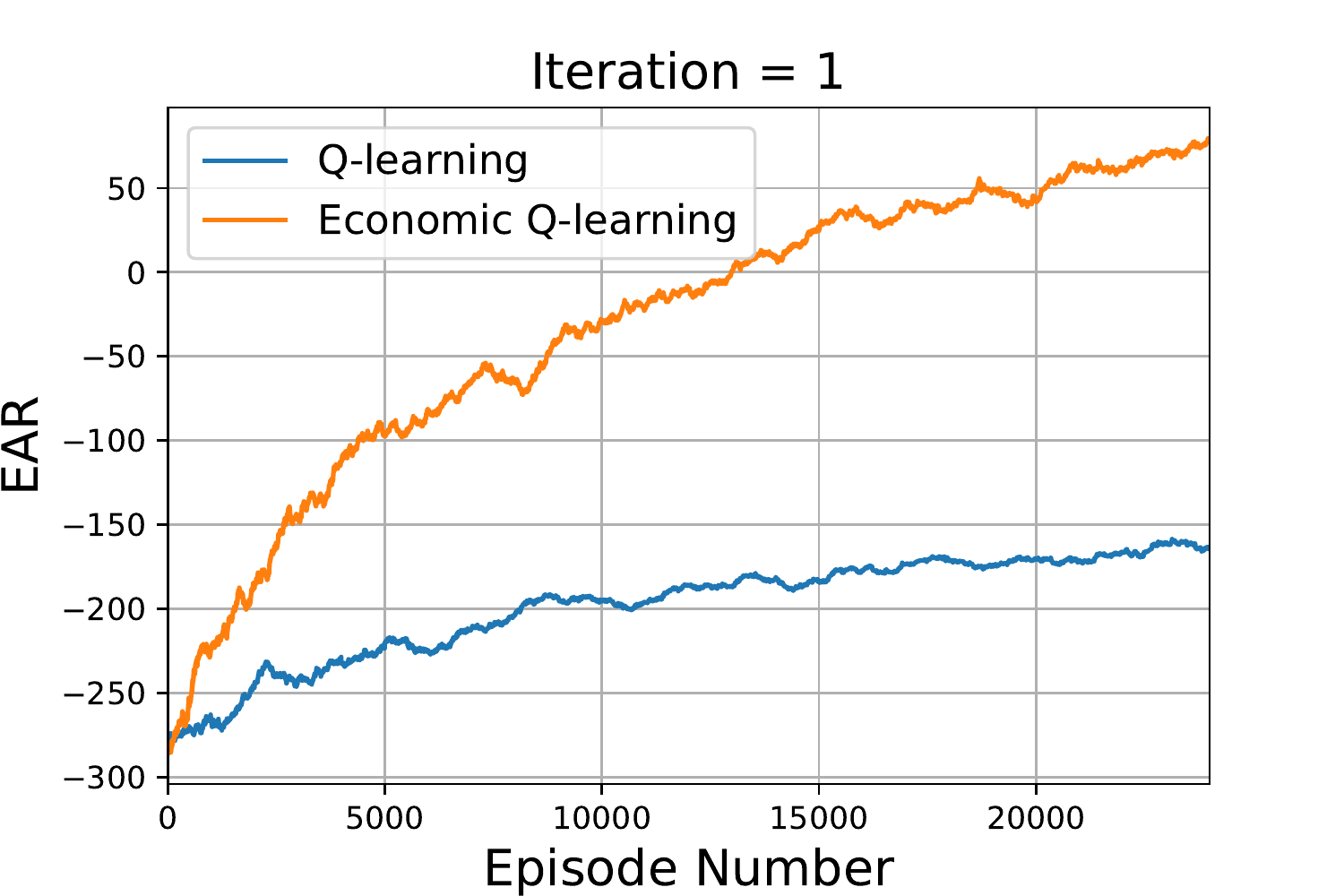}
        }     
        \hspace{-12pt} 
        \subfigure[]
        {
        \label{fig:reward2}
            \includegraphics[width=0.32\textwidth, keepaspectratio=true]{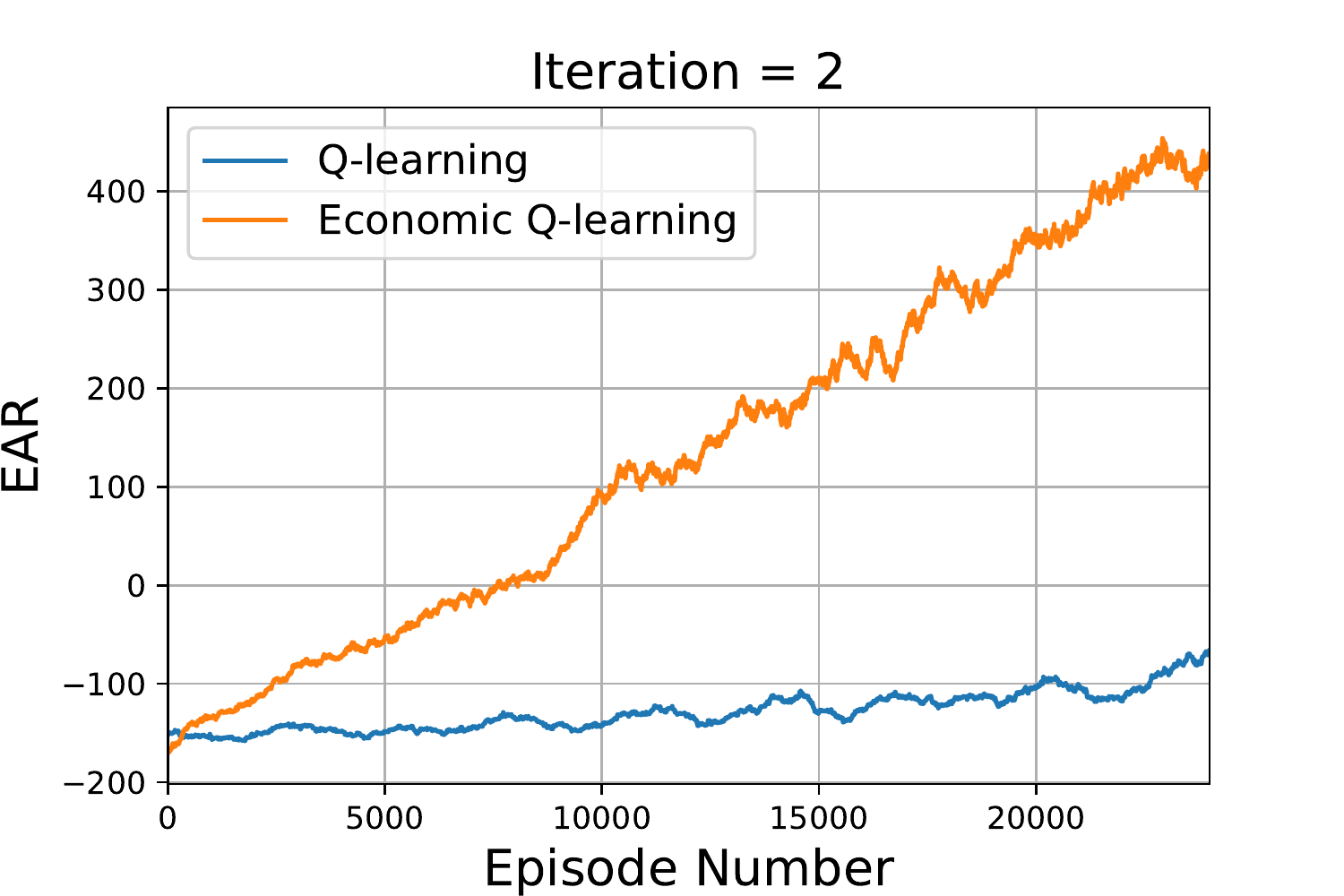}
        }        
        \hspace{-12pt} 
        \subfigure[]
        {
        \label{fig:reward3}
            \includegraphics[width=0.32\textwidth, keepaspectratio=true]{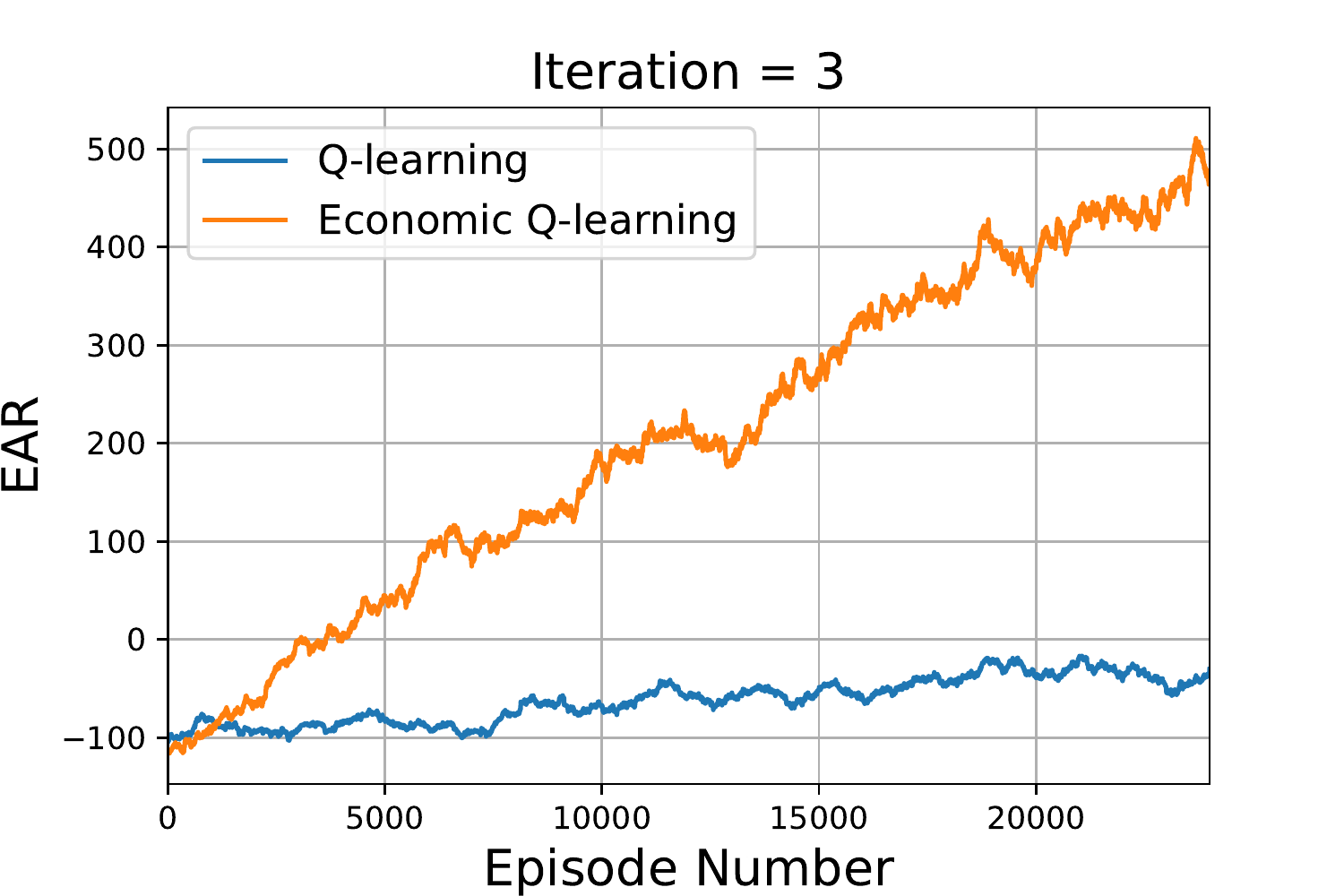}
        }        
    \end{center}
    \vspace{-15pt}
    \caption{EAR incurred by 3 UAV agents to visit 20 POIs in 24000 episodes after (a) 1 training iteration, (b) 2 training iterations, and (c) 3 training iterations, with Q-learning and economic Q-learning.}
    \label{fig:reward}
    \vspace{-9pt}
\end{figure*}


\section{Evaluation Result and Discussion}
\label{sec:evaluation_result}

In this section, we evaluate and analyze the \name{}’s performance through four different aspects. Firstly, we analyze the performance of the framework with respect to time for both economic and non-economic Q-learning. Then, we analyzed the efficiency of the model by observing the percentage of task completion for different sizes of UAV swarm (In this case, we kept the training episodes number fixed). Later, we show the performance of \name{} and non-economic Q-learning with respect to the total distance travelled to complete the goal set with different number POIs to visit. Lastly, we compare the reward gain by the economic and non-economic Q-learning agents through the training episodes in a iteration.

\subsection{Time for Completing Goal Set}
In this part, we specifically focus on comparing the economic and non-economic reinforcement learning with respect to the time required for completing the goal of the whole UAV swarm. We try with different number of episodes in a training iteration and show the trend for changing required time for both the economic and non-economic version of Q-learning. From Fig.~\ref{fig:time}, it is evident that economic Q-learning is performing better for each of the swarm size and remains better than the normal Q-learning with increasing training episodes. In Fig.~\ref{fig:time1}, it is seen that \name{} performs exceptionally well with each agent having larger goal set, that is, way more POIs to visit. Another point to be noticed is, with 200 and 300 training episodes in Fig.~\ref{fig:time2} and Fig.~\ref{fig:time3}, the Q-learning is getting close to the economic one, but still it is significant less efficient than than the economic Q-leraning.

\subsection{Percentage of Completion for Different Swarm Sizes}
In this part, we evaluate the performance of the \name{} model by observing the percentage of completion of the tasks. We necessarily keep the number of training episodes constant, which is 1000, and try with different number of UAVs. We calculate the completion of tasks with respect to steps needed and show how larger swarms can complete the task set faster with \name{} framework. Also, we try with different number of points to visit, to see how the swarm behavior changes with larger goal set. Fig.~\ref{fig:percent} represents the evaluations graphs for this metric. From Fig.~\ref{fig:percent1}, it is seen that after 75 steps, swarm size of 9 almost completed 90 percent of the goal set, where swarm size of 6 and 3 completed close to 70 and 55 percent respectively. As per Fig.~\ref{fig:percent2} and Fig.~\ref{fig:percent3}, it can be observed that with the increase of points in the environment, completion percentage goes down for swarm size of 3 and 6 and it requires more steps for swarm size of 9 to complete the goal set, which is reasonable. Also, in Fig.~\ref{fig:percent2}, it is seen that during steps 75 through 125, swarm size got really close to the efficiency of swarm size of 9, most likely due to the exploration behavior. As the number of points increased, swarm performance decreased, likely due to the extra processing time for point assignment. 

\subsection{Optimal Travelling Distance for completing the goal set}
Through Fig.~\ref{fig:distance}, we estimate the efficiency of the \name{} framework in optimizing the travelling distance for completing the goal set. Ideally, the more efficient model will require lesser distance to travel to complete the goal set, effectively optimizing the usage of resources and time. Here, we compare the Q-learning and economic Q-learning with different number of points to visit and show the distance they require to visit those points. Additionally, we try with different number of UAV swarm size in Fig.~\ref{fig:distance1}, Fig.~\ref{fig:distance2}, Fig.~\ref{fig:distance3} and observe the behavior alterations. 
Although, for all the different swarm sizes, economic Q-learning is doing better than the non-economic one, an improvement trend can be seen for the non-economic Q-learning with larger swarm size. From this behaviour, it can be inferred that the \name{} performs way better than the non-economic one with significantly larger number of points to visit.

\subsection{Reward Accumulation Visuals}
Finally, we evaluate the efficiency of the \name{} framework by comparing the achieved reward of the economic Q-learning with achieved reward of the regular Q-learning. This metric is the most theoretically important. Reinforcement learning algorithms optimize for the maximum reward. If a reward function is well designed, and balances different priorities appropriately, it also optimizes for environment behaviour. Despite the fact that our algorithm out-preforms the standard in UAV specific metrics, reward Accumulation shows the difference in optimizing power of the two algorithms. As such, these are the most impressive results of our work, providing evidence that performance can be increased even further and can provide benefits to other robotic control applications using reinforcement learning. We calculate the average reward gained by all the agents in the swarm in a particular episode to visualize this metric. In Fig.~\ref{fig:reward}, we see that both the models improve with respect iteration number. From Fig.~\ref{fig:reward1}, it is seen that both models started at around EAR of -280 and got better over the episodes. However, the economic Q-learning got better really fast and achieved an EAR of almost 75 at end of the first iteration, on 24000th episode, whereas the non-economic one barely reached an EAR of -150. In Fig.~\ref{fig:reward2}, at the beginning of second iteration, with the already updated Q-table, both models start around EAR of -150 and the same trend is followed as iteration 1. The interesting pattern is observed in Fig.~\ref{fig:reward3} for the third iteration, where we see that there is not much improvement for the economic Q-learning even after it started at an EAR of -100, which is higher than iteration 3. This behavior is expected because the total achievable reward in the environment is fixed and the economic model is already reaching close to that in the second iteration. Also, the improvement of the non-economic model is almost flat compared to the economic Q-learning.



\section{Conclusion}
\label{sec:conclusion}

In this paper, we have introduced a reinforcement learning algorithm coupled with an economic trading theory to distribute points and create paths for a multi-agent UAV surveillance problem. The proposed algorithm allows agents to negotiate flight plans with other agents via auction broadcasts and exchanging of POI contracts. Our evaluation results have shown that our approach outperforms the standard Q-learning method, allowing for UAVs to delegate tasks to other agents, and learn cooperative strategies in a multi-agent context. More specifically, with as few as three UAVs, we have observed a significant decrease of 33\% in the distance needed to travel to all POIs. This trend has appeared to be a function of the ratio of UAVs to POIs. As the number of POIs has increased relative to that of UAVs, the economic algorithm showed increasingly efficient paths. Furthermore, at every stage of training with any number of UAVs, the economic algorithm has been able to complete the swarm goal set faster. Additionally, the overall performance has continued to scale as UAVs are added to the swarm, the best shown by a 200\% greater reward that the economic Q-learning has accrued at the end of 1 training iteration using 3 UAVs. 
The UAVs have found efficient paths in environments with randomly initialized POI location without knowledge of other UAVs paths, conditioned solely on other UAV's instantaneous position. In addition, our results have shown that the addition of economic exchange allowed the UAVs to distribute the computation of effective paths through local decision making while ultimately optimizing for the global goal implicit in the reward function. The UAVs have been able to avoid collisions with the swarm's path plans. Through the auctions, the UAVs have avoided unnecessary competition and been able to focus on sections of the map. This feature shows the swarm strategizing as a whole without direct communication, with the auction operating over the Q-learning algorithm. Some pathological strategies have been learned by both economic and standard  Q-learning models. As such, our future works will focus on the refinement of the training process and the design of the reward function. The generality of the environment as a stochastic game allows for an extension of this method to other multi-agent problems. In addition, we will focus on the use of deep reinforcement learning to enable the integration of more detailed simulations and further refinement of the auction process.

\bibliographystyle{unsrt}
\bibliography{References}


\end{document}